\documentclass{article}

\usepackage{PRIMEarxiv}

\usepackage[utf8]{inputenc} 
\usepackage[T1]{fontenc}    
\usepackage{hyperref}       
\usepackage{url}            
\usepackage{booktabs}       
\usepackage{amsfonts}       
\usepackage{nicefrac}       
\usepackage{microtype}      
\usepackage{lipsum}
\usepackage{fancyhdr}       
\usepackage{graphicx}       
\usepackage{float}
\usepackage{placeins}
\usepackage[edges]{forest}
\usepackage{xcolor}
\usepackage{multirow}   
\usepackage{pifont}     
\usepackage{array}      
\graphicspath{{media/}}     
\graphicspath{{./}{figures/}{media/}}
\usepackage[acronym]{glossaries}
\newacronym{dl}{DL}{deep learning}
\newacronym{ai}{AI}{Artificial Intelligence}
\newacronym{ssl}{SSL}{Self-Supervised Learning}
\newacronym{mae}{MAE}{Masked Autoencoders}
\newacronym{cnn}{CNN}{Convolutional Neural Network}
\newacronym{vit}{ViT}{Vision Transformer}
\newacronym{mri}{MRI}{Magnetic Resonance Imaging}
\newacronym{ct}{CT}{Computed Tomography}
\newacronym{pet}{PET}{Positron Emission Tomography}
\newacronym{mssl}{MSSL}{Multimodal Self-Supervised Learning}
\newacronym{mim}{MIM}{Masked Image Modeling}
\newacronym{prisma}{PRISMA}{Preferred Reporting Items for Systematic Reviews and Meta-Analyses}
\newacronym{pgl}{PGL}{Prior-Guided Local}
\newacronym{drr}{DRR}{Digitally Reconstructed Radiograph}
\newacronym{nnclr}{NNCLR}{Nearest-Neighbor Contrastive Learning}
\newacronym{oct}{OCT}{Optical Coherence Tomography}
\newacronym{cloob}{CLOOB}{Contrastive Leave-One-Out-Boost}
\newacronym{clip}{CLIP}{Contrastive Language-Image Pre-training}
\newacronym{gan}{GAN}{Generative Adversarial Network}
\newacronym{auc}{AUC}{Area Under Curve}
\newacronym{ml}{ML}{Machine Learning}
\newacronym{gpu}{GPU}{Graphics Processing Unit}
\newacronym{jepa}{JEPA}{Joint-Embedding Predictive Architecture}

\usepackage{pgfplots}
\pgfplotsset{compat=1.18}
\usepackage{tikz}
\usetikzlibrary{shapes.geometric, arrows}
\usetikzlibrary{positioning, arrows.meta}
\title{Task-Aligned Self-Supervised Learning for Medical Image Analysis: A Task-Oriented Review with Practical Design Guidelines}

\author{
Chathura Wimalasiri\textsuperscript{} \quad
Yuchong Yao\textsuperscript{} \quad
Kishor Nandakishor\textsuperscript{} \quad
Marimuthu Palaniswami\textsuperscript{} \\[0.5em]
\textsuperscript{}Department of Electrical and Electronic Engineering, 
University of Melbourne, Parkville, VIC 3052, Australia \\[0.5em]
\texttt{c.kanakkahewage@student.unimelb.edu.au} \\
\texttt{yuchong.yao2@unimelb.edu.au, nandakishor.desai@unimelb.edu.au, palani@unimelb.edu.au}
}

\begin{document}
\maketitle

\begin{abstract}
Self-supervised learning (SSL) is increasingly used in medical image analysis to reduce dependence on costly expert annotations by learning transferable representations from unlabeled data. However, SSL performance depends not only on model architecture but also on whether the self-supervised objective preserves the information required by the downstream clinical task. This review presents a task-oriented synthesis of SSL methods for medical imaging, focusing on how the design of the self-supervised objective interacts with imaging modality, label availability, and downstream performance. We analyze $78$ studies published from 2017 to 2025 and organize them into four paradigms: contrastive, non-contrastive and predictive, generative and reconstruction-based, and hybrid learning. Rather than cataloging methods chronologically, we examine how these paradigms support classification, segmentation, detection, reconstruction, and regression. The evidence suggests that effectiveness is governed by the match among objective, modality, and downstream task rather than by any single strategy. Contrastive objectives favor global discriminative representations suited to classification but may underrepresent localized pathology, whereas spatial-prediction, masked-modeling, and reconstruction objectives better preserve anatomical structure for segmentation and dense prediction. Critically, misaligned objectives can cause negative transfer through shortcut learning on acquisition signatures or augmentation that erases diagnostic signal rather than merely weaker gains. SSL is most beneficial in low-label regimes, but its effectiveness depends on modality-aware augmentation, pathology-preserving corruption, and clinically meaningful evaluation. We conclude with practical design guidelines and open challenges for clinically aligned SSL.
\end{abstract}


\section{Introduction}
\label{sec:introduction}

\begin{figure}[ht]
    \centerline{\includegraphics[width=0.5\linewidth]{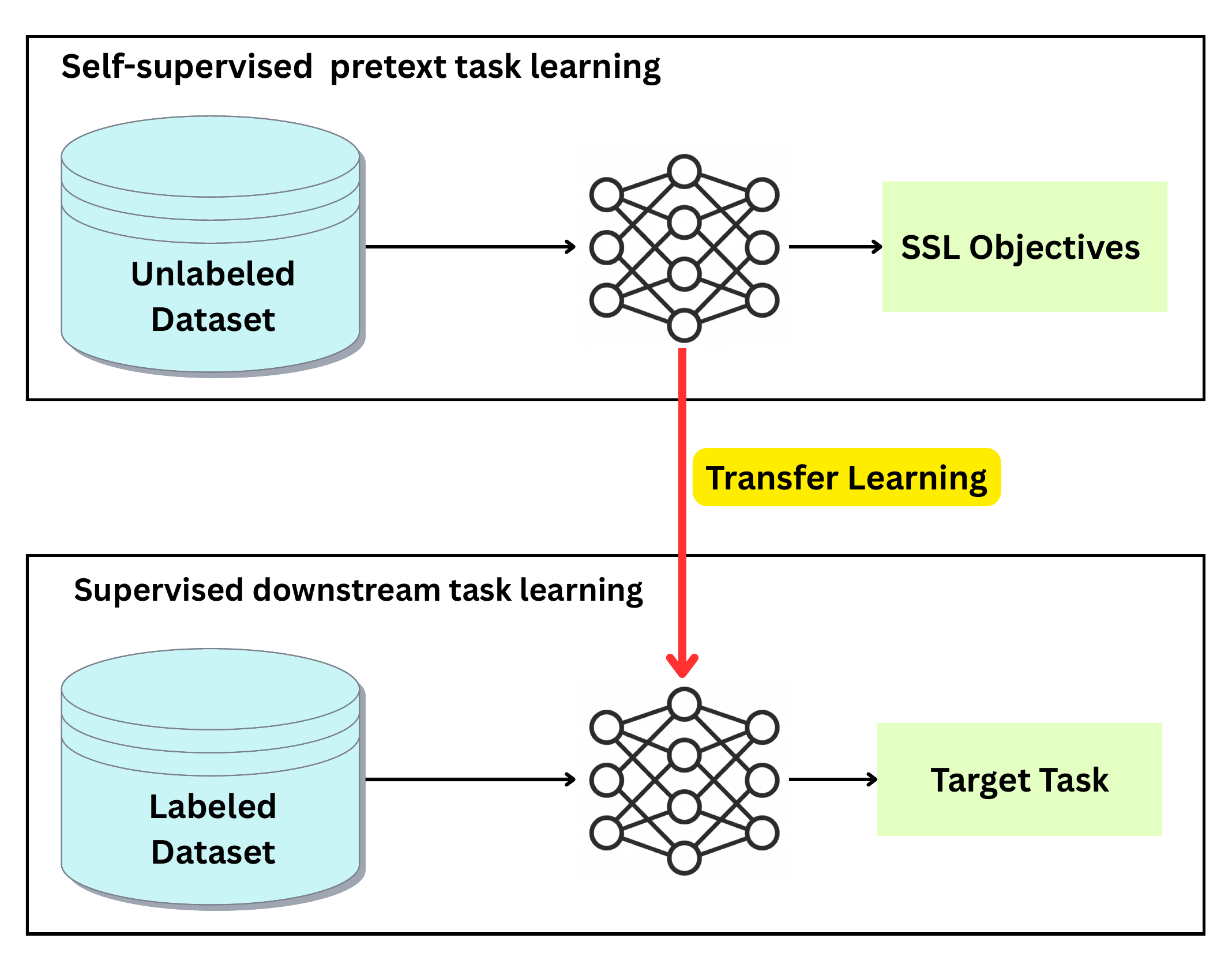}}
    \caption{General workflow of \gls{ssl}. The top panel illustrates the pretraining stage using unlabeled data, while the bottom panel shows adaptation to downstream tasks. Red arrows indicate the transfer learning process.}
    \label{figure1}
\end{figure}

\Gls{ssl} has become an important strategy in medical image analysis, particularly because expert annotations are expensive, time-consuming, and often require specialized clinical knowledge~\cite{willemink2020preparing, tajbakhsh2020embracing, shurrab2022self}. Unlike fully supervised learning, \gls{ssl} derives supervisory signals directly from unlabeled data and learns transferable representations through pretext tasks~\cite{jing2020self}. These learned representations can then be adapted to downstream clinical tasks such as disease classification, anatomical segmentation, lesion detection, image reconstruction, and prognosis prediction. Figure~\ref{figure1} illustrates this general \gls{ssl} workflow, in which a model is first pretrained on unlabeled data using a pretext objective and then transferred to a supervised downstream task.

In \gls{ssl}, ``pretext tasks" are designed to encourage models to learn useful visual representations without manual labels. Common examples include predicting missing image regions, recovering corrupted inputs, solving spatial arrangement tasks, or contrasting different augmented views of the same image~\cite{zhou2021review}. With the transition from \glspl{cnn} to \glspl{vit}, \gls{ssl} has become increasingly influential in medical \gls{ai}. Methods such as \gls{mae}~\cite{he2022masked} and contrastive learning~\cite{chen2020simple} have shown strong performance, especially in low-label settings where annotated medical images are limited.

Despite this progress, an important challenge remains: the effectiveness of \gls{ssl} depends not only on the architecture or the amount of unlabeled data, but also on how well the pretext task aligns with the downstream clinical objective. Early \gls{ssl} approaches in medical imaging often adopted pretext tasks from natural-image computer vision, such as rotation prediction~\cite{gidaris2018unsupervised} and jigsaw puzzle solving~\cite{noroozi2016unsupervised}. Although these tasks can improve general representation learning, they may not always preserve the clinically relevant information required for medical decision-making. For example, enforcing rotation invariance may be useful for some pathology or dermatology images, where orientation is less diagnostically meaningful, but it may be harmful in cardiac \gls{mri} or other anatomy-sensitive tasks where orientation and spatial relationships carry clinical significance.

This issue can be described as a ``\textbf{task-alignment problem}". There is no universal pretext task that is optimal for all medical imaging applications. Instead, the value of an \gls{ssl} representation depends on whether the pretext task preserves the type of information required by the downstream task~\cite{purushwalkam2020demystifying}. Clinical tasks impose overlapping representational demands: classification may depend on both global context and small localized abnormalities, while segmentation and detection require local precision together with broader anatomical context. A pretext task that removes or suppresses subtle lesions, orientation cues, or modality-specific contrast may perform well on generic benchmarks but fail in clinically important scenarios. Therefore, \gls{ssl} methods for medical imaging should be evaluated not only by average downstream performance, but also by their ability to preserve diagnostically relevant information.

Medical image analysis also differs from general computer vision because of the diversity and structure of imaging modalities. \gls{ct} reflects tissue density, \gls{mri} provides rich soft-tissue contrast, \gls{pet} captures metabolic activity, ultrasound contains temporal and operator-dependent information, and digital pathology includes high-resolution morphology and staining patterns~\cite{chartsias2017multimodal}. Even within a single modality, such as \gls{mri}, different sequences including T1, T2, and FLAIR provide complementary views of the same anatomy. This creates opportunities for both unimodal and cross-modal \gls{ssl}. For example, predicting one \gls{mri} sequence from another can encourage a model to learn anatomical representations that are robust to contrast differences~\cite{armanious2020medgan}. Similarly, aligning fundus photography with \gls{oct} or fluorescein angiography can help models capture shared disease-relevant features across retinal imaging modalities. These examples show that modality-aware pretext task design is central to effective \gls{ssl} in medical imaging~\cite{shurrab2022self}.

This review provides a task-oriented analysis of \gls{ssl} methods in medical image analysis, with a focus on the relationship between \gls{ssl} objective design and downstream clinical utility. We consider both unimodal \gls{ssl}, where representation learning is performed within a single imaging modality, and cross-modal \gls{ssl}, where multiple imaging modalities or views are used to generate supervisory signals. The review covers major medical imaging domains, including \gls{mri}, \gls{ct}, X-ray, ultrasound, retinal imaging, digital pathology, and other clinically relevant visual data.

Previous reviews have established the importance of \gls{ssl} in medical imaging, but they have approached method selection from different perspectives. Shurrab and Duwairi provided a broad taxonomy of \gls{ssl} methods and applications, whereas Huang et al.\ systematically reviewed \gls{ssl} for medical-image classification and proposed implementation guidance for that task \cite{shurrab2022self,huang2023ssl}. Zhang et al.\ examined pretext-task choice, architecture, class imbalance, and downstream training policies through controlled experiments, while VanBerlo et al.\ focused primarily on diagnostic classification and segmentation in X-ray, \gls{ct}, \gls{mri}, and ultrasound \cite{zhang2023dive,vanberlo2024survey}. Other recent reviews have cataloged \gls{ssl} applications across modalities or positioned self-supervision within the broader landscape of learning with limited annotations \cite{zeng2024self,kumari2026learning}. These studies provide important taxonomies, empirical comparisons, and practical recommendations. However, their principal organizing frameworks generally emphasize method families, imaging modalities, or downstream task categories. Comparatively less attention has been given to how the learning signal imposed by a pretext objective interacts with the information requirements of a particular clinical application. The present review addresses this issue through a task-alignment perspective that examines whether a pretext objective is likely to retain, emphasize, reconstruct, or reduce sensitivity to clinically relevant features, including localized pathology, anatomical organization, modality-specific contrast, temporal structure, and cross-modal correspondence. This analysis is interpretive and evidence-based rather than a formal measurement of information preservation. The notion of visual-information preservation has been used before at the level of a single training framework---most directly by PCRLv2~\cite{zhou2023pcrlv2}, which builds multi-scale pixel, semantic, and scale preservation into one model's pretraining objective. Our use of the idea is different in kind: rather than proposing a method that preserves information, we apply information preservation as an analytical lens across the reviewed studies, asking for each objective family which clinically relevant information it preserves, reconstructs, suppresses, or becomes invariant to, and how that predicts transfer under different modalities, pathology scales, and label regimes.

\begin{table*}[t]
    \centering
    \caption{Positioning of the present review relative to representative reviews and related empirical analyzes of self-supervised and label-efficient learning in medical imaging.}
    \label{tab:surveys}
    \footnotesize
    \setlength{\tabcolsep}{2pt}
    \resizebox{\textwidth}{!}{%
    \begin{tabular}{@{}p{2.3cm}p{2.6cm}p{3.5cm}p{4.5cm}p{4.5cm}@{}}
    \toprule
    \textbf{Review} & \textbf{Primary scope} & \textbf{Main organizing axis} 
    & \textbf{Downstream-task coverage and practical output} 
    & \textbf{Distinction from the present review} \\
    \midrule
    Shurrab and Duwairi~\cite{shurrab2022self} 
    & General \gls{ssl} methods and medical-imaging applications 
    & \gls{ssl} methodology and application domain 
    & Multiple medical-imaging tasks and modalities; broad methodological overview 
    & Does not organize the evidence primarily around the relationship between pretext-task learning signals and downstream representational requirements. \\
    Huang et al.~\cite{huang2023ssl} 
    & \gls{ssl} for medical-image classification 
    & \gls{ssl} strategy, modality, clinical domain, and fine-tuning procedure 
    & Classification-focused evidence synthesis and implementation guidance 
    & Its principal evidence synthesis focuses on classification rather than dense prediction, reconstruction, detection, and regression. \\
    Zhang et al.~\cite{zhang2023dive} 
    & Experimental evaluation of \gls{ssl} design choices 
    & Pretext tasks, architecture, class imbalance, and training policies 
    & Controlled benchmarking and empirical implementation guidance 
    & Provides controlled experiments rather than a cross-study synthesis across clinical modalities and downstream objectives. \\
    VanBerlo et al.~\cite{vanberlo2024survey} 
    & \gls{ssl} pretraining in radiological imaging 
    & Imaging modality and comparison with supervised learning 
    & Primarily diagnostic classification and segmentation in X-ray, \gls{ct}, \gls{mri}, and ultrasound; practitioner-oriented recommendations 
    & Does not formalize task alignment through the information-preservation requirements of individual downstream objectives. \\
    Zeng et al.~\cite{zeng2024self} 
    & Broad \gls{ssl} applications in medical imaging 
    & Imaging modality, \gls{ssl} family, and clinical application 
    & Classification, localization, segmentation, image-quality assessment, and related tasks; broad conceptual guidance 
    & Catalogs application patterns but does not systematically audit positive, limited, and negative transfer at the pretext--objective level. \\
    Kumari and Singh~\cite{kumari2026learning} 
    & Learning under varying supervision levels 
    & Supervision regime 
    & Broad medical-imaging tasks and label-efficient learning paradigms 
    & \gls{ssl} is considered within a broader supervision landscape rather than through a dedicated pretext--task alignment analysis. \\
    \textbf{Present review} 
    & \textbf{Task-aligned \gls{ssl} across major medical-imaging modalities} 
    & \textbf{Pretext learning signal $\times$ modality information $\times$ downstream representational demand} 
    & \textbf{Classification, segmentation, detection and anomaly localization, reconstruction, and regression. Qualitative task-alignment synthesis and conditional design guidance} 
    & \textbf{Interprets how the learning signals of different pretext objectives relate to downstream information requirements under different modalities, pathology scales, and label regimes.} \\
    \bottomrule
    \end{tabular}%
    }
    \par\vspace{2pt}
    \scriptsize\raggedright
    The comparison summarizes the principal analytical focus of each review rather than implying that topics not listed were completely absent.
\end{table*}

The main contributions of this review are as follows:
\begin{itemize}
    \item We introduce a \emph{task-alignment} perspective for \gls{ssl} in medical imaging, in which a \gls{ssl} objective is judged not by generic benchmark performance but by whether it preserves the information a specific downstream task requires.
    \item We synthesize the included studies according to pretext objective, imaging modality, downstream task, label regime, and reported transfer behavior, while highlighting representative limitations and cases of limited or negative transfer.
    \item We consolidate these findings into actionable design guidelines linking each downstream objective, imaging modality, and annotation regime to a recommended class of \gls{ssl} objective, and we characterize the conditions under which \gls{ssl} yields \emph{negative} transfer in clinical settings.
\end{itemize}

The remainder of this paper is organized as follows. Section~\ref{sec:literature-selection-and-synthesis} describes the literature selection and synthesis. Section~\ref{sec:background} provides background on \gls{ssl} paradigms and downstream clinical tasks. Section~\ref{sec:taxonomy-of-pretext-tasks-in-medical-imaging} presents the taxonomy of \gls{ssl} objectives in medical imaging. Section~\ref{sec:task-alignment-analysis} analyzes task alignment, modality effects, label regimes, and transfer behavior. Section~\ref{sec:practical-design-guidelines} provides practical design guidelines. Section~\ref{sec:open-challenges-and-limitations} discusses open challenges and limitations. Section~\ref{sec:future-research-directions} outlines future research directions, and Section~\ref{sec:conclusion} concludes the review.

\section{Literature selection and synthesis}
\label{sec:literature-selection-and-synthesis}

This review was conducted as a narrative, task-oriented synthesis of \gls{ssl} methods for medical image analysis, in which studies were selected purposively to illustrate the relationship between \gls{ssl} objective design and downstream clinical objectives rather than through an exhaustive systematic protocol. Studies published between 2017 and 2025 were identified through searches conducted up to \textbf{December 2025} of four bibliographic databases: PubMed, IEEE Xplore, Scopus, and Web of Science supplemented by Google Scholar, using combinations of keywords such as ``self-supervised learning,'' ``medical imaging,'' ``contrastive learning,'' ``masked image modeling,'' ``pretext task,'' ``segmentation,'' ``classification,'' ``MRI,'' ``CT,'' ``X-ray,'' ``ultrasound,'' ``retinal imaging,'' and ``histopathology.'' ArXiv preprints indexed through Google Scholar were also considered where they introduced widely used or frequently cited methods, and are identified as preprints in the reference list.

\begin{figure}[htbp]
    \centering
    \resizebox{0.65\textwidth}{!}{%
    \begin{tikzpicture}[
        node distance=1.5cm and 2.5cm,
        box/.style={draw, rectangle, minimum width=4.5cm, minimum height=1cm,
                    text width=4.3cm, align=center, font=\small},
        arrow/.style={-Stealth, thick}
    ]
    \node (start) [box] {\textbf{Identification} \\[4pt] Records identified from databases and preprint servers: PubMed, IEEE Xplore, Scopus, Web of Science, arXiv \\ (n = 854)};
    \node (dup) [box, below=of start] {\textbf{Screening} \\[4pt] Records after duplicates removed \\ (n = 759)};
    \node (eligible) [box, below=of dup] {\textbf{Eligibility} \\[4pt] Full-text articles assessed \\ (n = 84)};
    \node (final) [box, below=of eligible] {\textbf{Included} \\[4pt] Studies included in the final synthesis \\ (n = 78)};
    \node (excl1) [box, right=2.5cm of dup] {Records excluded after title/abstract screening \\ (n = 675)};
    \node (excl2) [box, right=2.5cm of eligible] {Full-text articles excluded: \\ -- Not focused on medical imaging \\ -- Lacked a self-supervised component or clear validation \\ (n = 6)};
    \draw [arrow] (start) -- (dup);
    \draw [arrow] (dup) -- (eligible);
    \draw [arrow] (eligible) -- (final);
    \draw [arrow] (dup) -- (excl1);
    \draw [arrow] (eligible) -- (excl2);
    \end{tikzpicture}%
    }
    \caption{Flow of study identification, screening, and selection for this task-oriented review.}
    \label{study_selection}
\end{figure}
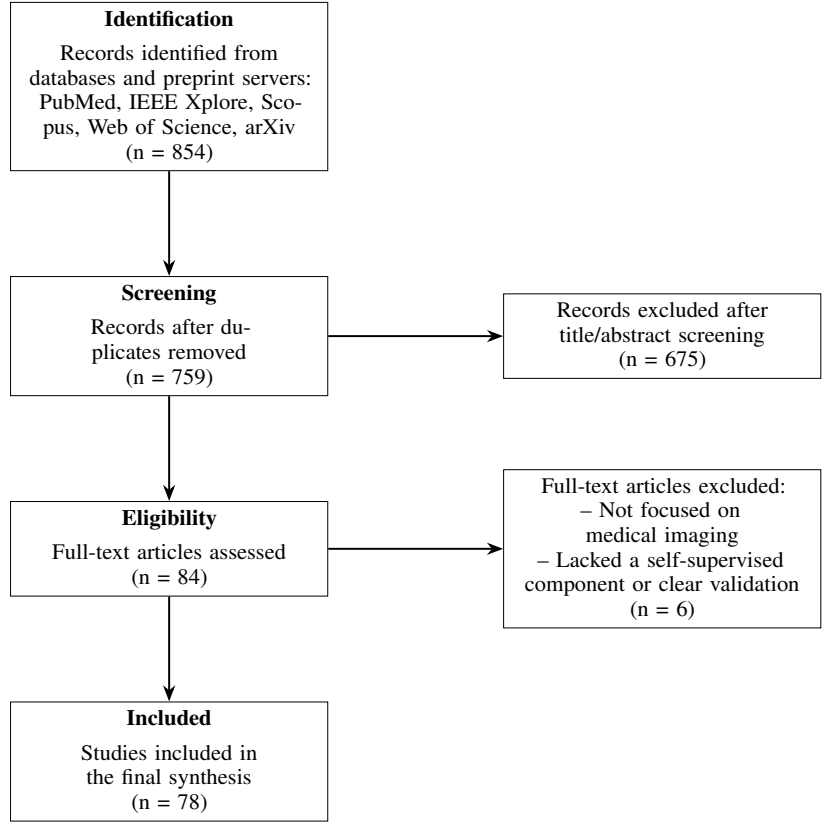

Studies were considered relevant if they applied or evaluated an \gls{ssl} method on medical imaging data and reported transfer to at least one downstream clinical task, such as classification, segmentation, detection, reconstruction, or regression. Studies focusing only on natural images, lacking a self-supervised component, or providing insufficient methodological detail were not considered. This criterion was applied equally to peer-reviewed articles and preprints. After applying these criteria, $78$ studies were retained for synthesis. Figure~\ref{study_selection} summarizes the identification, screening, and selection process.

Each study was analyzed according to its \gls{ssl} paradigm, \gls{ssl} objective design, imaging modality, downstream task, label regime, and reported transfer performance. Because the reviewed studies used heterogeneous datasets, architectures, and evaluation protocols, a quantitative meta-analysis was not performed. Instead, we provide a qualitative synthesis focused on task alignment, modality-aware design, label efficiency, and positive or negative transfer. The qualitative alignment ratings in Table~\ref{tab:alignment} compare reported transfer outcomes with the representational demands of each downstream task. Ratings were assigned by the lead author, reviewed by the co-authors, and resolved through reference to the primary studies. Ratings based on limited or indirect evidence are marked accordingly. As they summarize the direction and consistency of transfer rather than pooled effect sizes, they should be interpreted as design guidance, not a quantitative meta-analysis.

\section{Background}
\label{sec:background}


\subsection{Self-supervised learning in medical imaging}

To provide a comprehensive analysis of \gls{ssl} objective design, we first discuss the foundational taxonomies of \gls{ssl} and the specific characteristics of medical images. This background is essential for understanding the ``alignment" between the \gls{ssl} objectives and the downstream clinical application.

\gls{ssl} learns feature representations from unlabeled data that are useful for downstream tasks. The \gls{ssl} objective plays a central role in this process.

Historically, \gls{ssl} methods were often grouped into predictive, generative, and contrastive families~\cite{shurrab2022self}, although current methods also include negative-free and hybrid objectives.

\begin{enumerate}
    \item Predictive Learning : These approaches apply a deterministic transformation $T(\cdot)$ to input image, such as rotation prediction, jigsaw permutation, or relative patch positioning and train the network to classify which transformation was applied~\cite{taleb20203d, zhu2020rubik}.
    
    \item Generative Learning : These approaches predict missing or corrupted information using targets such as normalized pixels, discrete tokens, latent teacher features, frequency components, or mesh embeddings. Early methods used bottleneck autoencoding architectures to reconstruct inputs pixel-by-pixel~\cite{vincent2008extracting}. More recent approaches, such as \gls{mim}~\cite{he2022masked} mask a portion of the input patches and train the network to predict the masked or missing patches. 

    By predicting masked content, these models encourage learning global context, local structure, and modality-specific regularities without relying primarily on view-based augmentation. In medical imaging, this method extended to predict one modality from another modality, thereby learning modality-invariant anatomical representations~\cite{armanious2020medgan}.
    
    \item Contrastive Learning : These approaches operate in latent feature space (Ex: SimCLR~\cite{chen2020simple}, MoCo~\cite{he2020momentum}). The objective is to maximize the similarity between "positive pairs" (two augmented views of the same image) while minimizing the similarity between "negative pairs" (views of different images).

    Some instance-discrimination methods treat each image as a separate identity, whereas other contrastive and joint-embedding methods define similarity through patients, regions, modalities, or teacher targets. These objectives can produce transferable representations when their similarity assumptions and augmentations preserve clinically relevant information. To better support dense prediction tasks such as segmentation, recent work extends this method to pixel or patch level~\cite{wang2021dense}. This method keeps the consistency between corresponding local regions across augmented views. 
\end{enumerate}

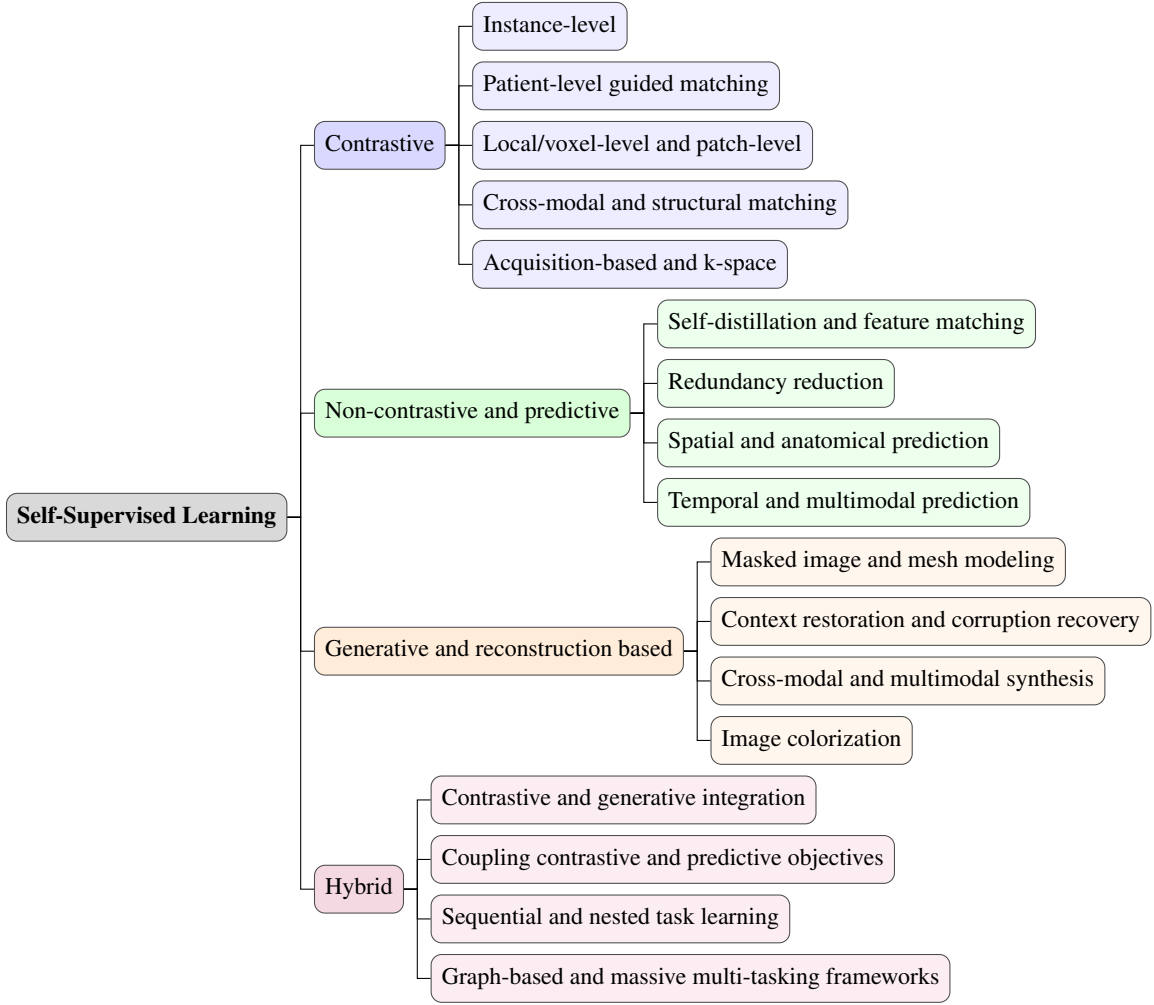
\begin{figure}[ht]
\centering
\begin{forest}
  forked edges,
  for tree={
    grow=east,
    reversed=true,
    anchor=base west,
    parent anchor=east,
    child anchor=west,
    base=left,
    font=\footnotesize,
    rectangle,
    draw=black!70,
    rounded corners,
    align=left,
    inner sep=4pt,
    s sep=4pt,
    l sep=10pt,
  },
  [Self-Supervised Learning, fill=gray!30, font=\footnotesize\bfseries
    [Contrastive, fill=blue!15
      [Instance-level, fill=blue!7]
      [Patient-level guided matching, fill=blue!7]
      [Local/voxel-level and patch-level, fill=blue!7]
      [Cross-modal and structural matching, fill=blue!7]
      [Acquisition-based and k-space, fill=blue!7]
    ]
    [Non-contrastive and predictive, fill=green!15
      [Self-distillation and feature matching, fill=green!7]
      [Redundancy reduction, fill=green!7]
      [Spatial and anatomical prediction, fill=green!7]
      [Temporal and multimodal prediction, fill=green!7]
    ]
    [Generative and reconstruction based, fill=orange!15
      [Masked image and mesh modeling, fill=orange!7]
      [Context restoration and corruption recovery, fill=orange!7]
      [Cross-modal and multimodal synthesis, fill=orange!7]
      [Image colorization, fill=orange!7]
    ]
    [Hybrid, fill=purple!15
      [Contrastive and generative integration, fill=purple!7]
      [Coupling contrastive and predictive objectives, fill=purple!7]
      [Sequential and nested task learning, fill=purple!7]
      [Graph-based and massive multi-tasking frameworks, fill=purple!7]
    ]
  ]
\end{forest}
\caption{Literature-derived taxonomy of \gls{ssl} objective strategies in medical image analysis.}
\label{figure2}
\end{figure}

Current frameworks also include negative-free representation learning, which avoids explicit negative pairs, and hybrid objectives that combine several learning signals~\cite{zhang2025survey, giakoumoglou2024review, zhao2025maemc}. Therefore, we organize self-supervised objectives into four groups: contrastive, non-contrastive and predictive, generative and reconstruction-based, and hybrid, as illustrated in Figure~\ref{figure2}. This categorization captures recent methodological developments and provides a structured foundation for the subsequent task-alignment analysis. Throughout this review, \emph{pretext task}, \emph{\gls{ssl} objective}, and \emph{learning signal} refer to the same underlying construct: the supervisory signal imposed during pretraining. We prefer \emph{objective} for families such as instance discrimination and self-distillation, which do not pose an explicit surrogate task.

\subsection{Downstream clinical tasks}

Pretrained \gls{ssl} models are evaluated by transferring them to downstream clinical tasks. Classification assigns an image or patient to a diagnostic category, such as disease screening or grading. Segmentation requires dense, pixel- or voxel-level prediction to delineate organs, lesions, or tissue boundaries. Detection and anomaly localization identify the presence and location of clinically relevant findings, including out-of-distribution or pathological regions. Reconstruction recovers high-fidelity images from incomplete inputs, as in undersampled \gls{mri}, while regression predicts continuous clinical quantities such as biomarker values, cellularity, or survival. These objectives place different demands on the learned representation: global semantic discrimination for classification, local structural fidelity for dense prediction. This is precisely why \gls{ssl} objective–task alignment matters, as the remainder of this review shows.

\subsection{Task alignment in self-supervised learning}

A central issue in medical \gls{ssl} is the alignment between the pretext task and the downstream clinical objective. Task alignment refers to the extent to which the information learned during \gls{ssl} pretraining is relevant to the target task. A well-aligned pretext task preserves clinically meaningful information, whereas a poorly aligned pretext task may encourage invariance that removes important diagnostic cues.

For example, rotation prediction or rotation-invariant contrastive learning may be useful in settings where orientation is not clinically meaningful, such as some histopathology or dermatology tasks. However, the same invariance may be harmful in anatomy-sensitive modalities such as cardiac \gls{mri}, where spatial orientation and anatomical arrangement are important. Similarly, aggressive masking may work well for natural images but may remove small lesions or subtle abnormalities in medical images.

Therefore, \gls{ssl} design in medical imaging should be task-aware and modality-aware. The choice of pretext task should depend on the downstream objective, the imaging modality, the expected pathology, and the available label regime. This review uses this task-alignment perspective to organize and analyze \gls{ssl} methods in medical image analysis.

\section{Taxonomy of SSL objectives in medical imaging}
\label{sec:taxonomy-of-pretext-tasks-in-medical-imaging}

Building on the four-family scheme motivated in Section~\ref{sec:background}, we organize the reviewed \gls{ssl} objectives into Contrastive; Non-contrastive and predictive; Generative and reconstruction-based; and Hybrid learning. The first three families are defined by their dominant \gls{ssl} objective, while Hybrid methods explicitly optimize objectives spanning more than one family. This taxonomy is structured by the dominant learning objective and by how supervision is derived from unlabeled data, providing the foundation for the downstream task-alignment analysis in Section~\ref{sec:task-alignment-analysis}.

\subsection{Contrastive learning}

\subsubsection{Instance-level contrastive learning}
\label{subsubsec:instance}

The standard contrastive formulation treats two augmented views of the same image as a positive pair, while views from different images serve as negatives. SimCLR~\cite{chen2020simple} and MoCo~\cite{he2020momentum} have been widely adapted to medical imaging across histopathology~\cite{ciga2022self}, ultrasound~\cite{ali2023self}, prostate \gls{mri}~\cite{de2025self}, and brain \gls{mri}~\cite{meng2024self}. Domain adaptations often lie in the augmentation policy rather than the loss: prostate \gls{mri} classification uses Rician noise and intensity shifts across T2W, DWI, and ADC sequences to learn robust slice-level features~\cite{de2025self}, while chest X-ray pretraining such as CheSS relies on standard MoCo augmentations at scale~\cite{cho2023chess}. The memory-bank formulation has been effective for COVID-19 prognosis and few-shot diagnosis~\cite{sriram2021covid, chen2021momentum}, and the \gls{nnclr} variant has been used for neuro-degenerative disorder classification on coronal T1 slices~\cite{gryshchuk2025contrastive}. A notable departure is PaRCL~\cite{yi2024parcl}, which constructs positive pairs by aggregating multi-layer feature representations of the same fundus image through a sub-network ensemble, avoiding strong augmentation entirely.

Instance discrimination encourages representations that are invariant to the chosen augmentations and separable at the level of individual images, and two consequences require care in medical imaging. First, the objective induces invariance to whatever the augmentation pipeline removes: aggressive random cropping can discard the small region carrying the diagnostic signal, and color or intensity jittering can distort quantitative information (Hounsfield values in \gls{ct}, standardized uptake in \gls{pet}, stain intensity in histopathology) that is itself label-relevant. Second, treating every image as its own class assumes different images are semantically distinct. In datasets with many scans of the same organ or the same normal anatomy, this manufactures false negatives that push biologically similar cases apart. Instance-level separability moreover does not entail clinical-class separability: nothing in the objective prevents the encoder from solving the task through scanner, institution, acquisition-protocol, or patient-identity cues, a shortcut that inflates pretext performance while degrading transfer. Reported outcomes are additionally sensitive to batch size, memory-queue length, and temperature, so gains should be read alongside the augmentation policy and the composition of the pretraining cohort.

\subsubsection{Patient-level guided matching} 
\label{subsubsec:patient}

Because multiple images of the same patient often share pathology-specific information, several works redefine positive pairs using patient identity. MICLe~\cite{azizi2021big} samples two images from a patient-level ``bag'' collected under different views or lighting. Self-FI adds bilateral matching between left and right eyes of the same patient~\cite{nguyen2023self}, while Jamaludin et al.\ pair vertebral bodies from longitudinal spinal \gls{mri} of the same patient and separate different individuals through a Siamese contrastive loss~\cite{jamaludin2017self}. A retinal framework by Li et al.\ extends this idea by matching a fundus image, its augmented view, and a synthesized fundus fluorescein angiography from the same patient~\cite{li2020self}.

These patient-level strategies are not equivalent, because each definition of a positive pair encodes a different assumption about what should remain invariant. MICLe~\cite{azizi2021big} treats differences in view and illumination within the same examination as nuisance variation. Self-FI~\cite{nguyen2023self} assumes symmetry between bilateral organs by matching the left and right eyes of the same patient. Repeated acquisitions of the same anatomical region similarly treat acquisition-related variation as irrelevant. Jamaludin et al.~\cite{jamaludin2017self} extend this assumption to longitudinal spinal \gls{mri}, treating changes between time points as nuisance variation. Li et al.~\cite{li2020self} instead assume that images from different modalities preserve shared patient-specific anatomy by matching fundus photographs with fluorescein angiography. Among these approaches, the bilateral and longitudinal variants may be particularly risky. Enforcing similarity between a diseased and a healthy eye, or between time points that straddle disease onset, may suppress precisely the pathological findings that a downstream task must detect.

\subsubsection{Local, voxel-level, and patch-level matching}

Dense prediction requires preserving spatial structure, which motivates localized contrastive objectives. vox2vec predicts whether two voxels from overlapping 3D patches correspond to the same anatomical location~\cite{goncharov2023vox2vec}. For 3D dental segmentation, Ma et al.\ use K-Nearest Neighbors on augmented dental meshes to sample overlapping jaw regions and match features across them~\cite{ma2025multi}. Liu et al.\ target 3D necrotic lung lesion segmentation with a distance-transform-based mask-out augmentation that simulates necrotic cores, producing slice-invariant and necrosis-robust features within RECIST-cropped regions~\cite{liu20243d}. Yan et al.\ combine instance-level patch contrast on WSIs with a clustering objective, applying contrastive loss in the soft K-way cluster-assignment space~\cite{yan2022deep}. VoCo predicts which anatomical region a randomly cropped sub-volume occupies by contrasting it against a grid of base crops, turning the consistent contextual position of organs in 3D scans into a contrastive position-prediction signal, and scales to large unlabeled \gls{ct} collections~\cite{wu2024voco}. At larger scale, frameworks that pair localized contrastive objectives with reconstruction---such as PCRLv2 and Swin UNETR---are discussed under hybrid learning (Section~\ref{subsec:hybrid}).

\subsubsection{Cross-modal and structural matching}
\label{subsubsec:crossmodal}

Contrastive learning extends beyond augmentation-based positives to cross-modal and structural relationships. In retinal imaging, \gls{clip}- and \gls{cloob}-style objectives align 2D fundus images with 3D \gls{oct} volumes from the same eye in a shared latent space~\cite{suekei2024multi}. Lung ultrasound frameworks treat M-mode slices extracted from the same B-mode video as positive pairs~\cite{vanberlo2023exploring}. LVM-Med formulates \gls{ssl} as second-order graph matching, using edge and neighborhood relationships between samples so that similar images remain similarly connected across augmented graphs~\cite{mh2023lvm}.

\subsubsection{Acquisition-based and k-space contrastive learning}
\label{subsubsec:kspace}

CL-\gls{mri} is unique to medical acquisition: instead of image-domain augmentations, it generates views by applying different undersampling operators and acceleration factors to the same k-space data, then maximizes mutual information across undersampling levels to produce an undersampling-invariant feature extractor for downstream reconstruction~\cite{ekanayake2025cl}.

\paragraph{Synthesis.} Contrastive objectives preserve globally discriminative, augmentation-invariant semantics, and induce invariance to whatever the augmentation set removes. Local and voxel-level variants additionally preserve spatial correspondence. They align with classification and, in localized form, with segmentation, but are weakly aligned with reconstruction, whose pixel-level detail their invariance discard. Characteristic failure modes are augmentation-induced loss of small pathology, false negatives among biologically similar cases, and shortcut learning on acquisition metadata. Evidence is strong and consistent for classification, moderate for dense prediction, and thin for the cross-modal and k-space variants.

\subsection{Non-contrastive and predictive learning}

Negative-free methods avoid explicit negative sampling, which can incorrectly separate biologically similar samples. Collapse is prevented through asymmetric predictors and stop-gradient operations in SimSiam-~\cite{chen2021simsiam} and BYOL-style~\cite{grill2020byol} methods, teacher--student centering and sharpening in DINO~\cite{caron2021dino}, and variance, covariance, or cross-correlation constraints in VICReg-~\cite{bardes2022vicreg} and Barlow Twins-style~\cite{zbontar2021barlow} objectives.

\subsubsection{Self-distillation and feature matching}

DINO-style~\cite{caron2021dino} teacher--student frameworks have been used for dental caries classification and external cervical resorption detection~\cite{mohammad2024artificial} and for periapical film segmentation in limited-label settings~\cite{hu2025self}. \gls{pgl} extends the idea to 3D volumes by matching voxel-level features of the same anatomical region across two transformed views rather than global embeddings~\cite{xie2020pgl}. Hirsch et al.\ adapt Masked Siamese Networks to endoscopic video, predicting prototype similarity distributions of an unmasked target view from a masked anchor view~\cite{hirsch2023self}. Vim4Path applies DINO-style~\cite{caron2021dino} distillation to histopathology using a Vision Mamba backbone, aligning global and local view distributions of whole-slide patches for slide-level classification via multiple instance learning~\cite{nasiri2024vim4path}. CADS extends self-distillation to cross-modal volumetric settings: a 2D \gls{drr} X-ray is generated from each 3D \gls{ct} volume, and a transformer-based student--EMA teacher aligns \gls{ct} and X-ray representations with bidirectional distillation across corrupted views and multiple encoder stages~\cite{ye2024cads}.

\subsubsection{Redundancy reduction}

Redundancy-reduction methods shape representation geometry through statistical constraints rather than pairwise discrimination. Barlow Twins~\cite{zbontar2021barlow} minimizes the deviation of the cross-correlation matrix between two augmented batches from the identity, and has been used for label-efficient dental caries classification~\cite{taleb2022self} and for high-resolution medical anomaly detection through a sliding-window strategy~\cite{dong2023swssl}.

\subsubsection{Spatial and anatomical prediction}
\label{subsubsec:spatial}

Rubik's Cube-based methods divide 3D scans into eight sub-volumes and require the model to predict the applied rotation and permutation~\cite{zhuang2019self}. Rubik's Cube+ adds cube masking identification~\cite{zhu2020rubik}, and volume-wise transformation strategies rotate entire anatomical layers along sagittal, coronal, or axial planes for restoration~\cite{tao2020revisiting}. Slice-level ordering has been used for body-part recognition on \gls{ct}~\cite{zhang2017self}, and spatial-awareness training inserts patches from neighboring slices and predicts their relative displacement~\cite{nguyen2020self}. Bai et al.\ overlay 2Ch and 4Ch \gls{mri} view planes to define nine anatomical regions and pretrain a pixel-wise classifier over them for cardiac segmentation~\cite{bai2019self}. In cytoarchitectonic analysis, Siamese networks predict the 3D geodesic distance between two small brain tissue patches~\cite{spitzer2018improving}.

More recent variants move from generic geometric transformations to domain-specific structural prediction. Med-SeAM pseudo-labels sagittal \gls{mri} slices as critical or non-critical based on anatomical position, then trains a CNN+BiGRU model with cross-attention to classify slice importance~\cite{daydar2024med}. The ASPECTS pretext task uses atlas-based registration to generate pseudo-labels for ten clinically defined brain regions on unlabeled \gls{ct} volumes, then pretrains a network to perform multiclass segmentation of these regions, encouraging hemisphere-specific anatomical and vascular-territory features directly relevant to stroke analysis~\cite{moreau2025self}.

It is useful to separate three levels within this group, because they draw on
increasingly strong external knowledge.

\emph{(i) Generic geometric prediction.} Rubik's Cube~\cite{zhuang2019self}, Rubik's Cube+~\cite{zhu2020rubik}, volume-wise rotation~\cite{tao2020revisiting}, slice ordering~\cite{zhang2017self}, and neighboring slice displacement~\cite{nguyen2020self} recover an applied geometric transform. The signal is intrinsic and modality agnostic; it teaches spatial consistency but carries no anatomical semantics and can be solved by low-level cues.

\emph{(ii) Intrinsic anatomical-position prediction.} Bai et al.~\cite{bai2019self} (cardiac view planes mapped to anatomical regions) and Spitzer et al.~\cite{spitzer2018improving} (3D geodesic distance between tissue patches) predict positions defined by the anatomy already present in the image, injecting anatomical structure without external labels. Zhang et al.~\cite{zhang2024anatomy} exploit anatomy-oriented imaging planes by regressing the relative orientation and location between planes, and report that transfer improves when the pretext target is anatomically relevant to the downstream task and direct support for the task-alignment premise of this review.

\emph{(iii) Atlas- or pseudo-label-guided anatomical supervision.} Med-SeAM~\cite{daydar2024med} and the ASPECTS pretext~\cite{moreau2025self} differ in kind: their targets are generated from external anatomical knowledge---atlas registration, clinically defined region boundaries---rather than from a transform the model itself applied. This places them closer to weak or surrogate supervision than to self-supervision in the strict sense, since the pretext labels encode prior clinical knowledge (vascular territories, region criticality). This is why their transfer to the matching clinical task is strong, and why comparing them directly to a rotation or slice-order pretext is not like-for-like.

\subsubsection{Temporal and multi-modal prediction}

Jiao et al.\ combine temporal order correction with spatio-temporal transform prediction (translation, scale, rotation, shear) for ultrasound video~\cite{jiao2020self}, and Manna et al.\ use jumbled patch arrangement prediction on knee \gls{mri}~\cite{manna2023self}. Taleb et al.\ extend spatial permutation prediction to a multi-modal setting: patches from different modalities are shuffled, and the model recovers the correct arrangement through a Sinkhorn-based permutation matrix~\cite{taleb2021multimodal}.

\paragraph{Synthesis.} This family preserves intrinsic structural and anatomical relations and, in its atlas-guided variants, injects external anatomical priors. Joint-embedding methods induce invariance without negatives, sidestepping the false-negative problem. Spatial and anatomical prediction align well with segmentation, and self-distillation transfers broadly. Generic geometric pretexts risk being solved by low-level shortcuts, and the strongest ``anatomical'' results draw partly on weak supervision rather than self-supervision alone, complicating attribution. Evidence is strong for segmentation and moderate elsewhere.

\subsection{Generative and reconstruction-based learning}

\subsubsection{Masked image and mesh modeling}
\label{subsubsec:mim}

An encoder processes a partially masked input, and a decoder reconstructs the missing information. \gls{mae} has been applied to COVID-19 chest X-ray classification with a Vision Transformer encoder--decoder~\cite{xing2023self}. For dental panoramic radiographs, Almalki and Latecki~\cite{almalki2023self} adapt SimMIM~\cite{xie2022simmim}, which replaces masked patches with learnable tokens inside the encoder to preserve positional structure and uses a lightweight linear head for pixel prediction. In histopathology, a pyramid-based masked image modeling framework combines a high masking ratio (75\%) with wavelet-based downsampling to capture both global tissue organization and local texture~\cite{wang2023pyramid}. SparK uses sparse convolutions on visible patches only and reconstructs masked regions, showing robustness in small-sample \gls{ct} regimes~\cite{wolf2023self}. Kidney tumor classification on \gls{ct} pairs masked-autoencoder pretraining with a self-distillation mechanism that transfers global semantic information from decoder to encoder~\cite{ozbay2024kidney}. DentalMAE extends masked modeling to non-grid 3D intra-oral scans by masking non-overlapping mesh face patches and reconstructing their embeddings rather than raw geometry~\cite{almalki2024self}.

A recurring claim in medical \gls{mim} is that masking ratios should be lower than the $60$--$75\%$ natural-image default because small pathological structures are easily lost. The reviewed evidence supports this only conditionally. It holds where the finding is small, sparse, and localized (pulmonary nodules, dental restorations), where high masking can remove the target region outright~\cite{zhao2025maemc, almalki2023self}. It does not generalize: a large controlled study of 3D MAE pretraining on brain \gls{mri} found static ratios of $60\%$ and $75\%$ and a dynamic $60$--$90\%$ ratio to perform equally well, and adopted the high dynamic ratio as its default~\cite{wald2025revisiting}. The optimal ratio is therefore task-, architecture-, and reconstruction-target-dependent, governed by the spatial scale of the diagnostic evidence and the density of anatomical information, rather than by a fixed ``lower-is-better'' rule.

\subsubsection{Context restoration and corruption recovery}

Some generative approaches apply physical or structural corruptions and train the network to restore the original image. Models Genesis unifies nonlinear intensity distortion, local pixel shuffling, and in/out-painting into a single 3D \gls{ct} restoration objective~\cite{zhou2019models}. A physics-inspired framework for perfusion estimation recovers unaugmented lung-cropped 3D \gls{ct} volumes from inputs corrupted by blur, noise, and elastic transformations to capture robust HU-density and inhale--exhale cues~\cite{liu2025perfusion}. Context restoration via patch swapping disrupts spatial context while preserving intensity distribution, and has been applied across ultrasound, \gls{ct}, and \gls{mri}~\cite{chen2019self}. Inpainting-based \gls{ssl} evaluates zeroing and swapping variants of this idea~\cite{dominic2023improving}. For anomaly detection, iterative latent token masking models discrete tokens with a VQ-VAE and autoregressive Transformer, iteratively masking low-likelihood tokens so that reconstruction learns a normal-data distribution even when training data contain anomalies~\cite{patel2023self}. Tahghighi et al.\ apply a masking-based residual reconstruction objective (optimal mask ratio 0.4) to 3D \gls{mri} volumes, with reconstruction and anti-symmetry constraints applied only within masked brain-surface cuboids~\cite{tahghighi2024enhancing}.

\subsubsection{Cross-modal and multi-modal synthesis}

In ophthalmology, several works reconstruct fluorescein angiography regions of interest from retinography inputs to learn shared and modality-specific retinal representations~\cite{hervella2020learning, morano2020multimodal, hervella2021self}. Multimodal Image Encoding further disentangles common and exclusive feature branches during cross-modal reconstruction~\cite{hervella2022multimodal}. Auto-GAN~\cite{cao2020auto} addresses missing \gls{mri} sequences: a collaborative \gls{gan} generator synthesizes a missing modality (e.g., T2-FLAIR from T1, T1-C, and T2), with target-modality self-reconstruction providing feature-level constraints.

\subsubsection{Image colorization}

Colorization is uncommon in medical imaging because most modalities are grayscale. In endoscopic video, a conditional \gls{gan} predicts $a,b$ chroma channels from $L$ luminance in Lab space~\cite{ross2018exploiting}. Self-HER2Net extends colorization to a multi-task generative scheme for breast cancer histopathology, transforming RGB IHC patches into HSL, HSV, a DAB stain space via color deconvolution, and a synthetic Hematoxylin-only stain to learn morphology and staining patterns critical for HER2 scoring~\cite{chyrmang2025self}.

\paragraph{Synthesis.} Reconstruction objectives preserve fine-grained local texture and spatial detail, making them the best-aligned family for dense prediction, detection, and reconstruction. The reconstruction target (pixels vs.\ discrete tokens vs.\ latent features) shifts the balance between low-level fidelity and semantic content. The dominant failure mode is corruption that destroys the diagnostic signal, so corruption and target design must be modality-aware. Evidence is strong for segmentation and reconstruction and growing for anomaly detection.

\subsection{Hybrid learning}
\label{subsec:hybrid}

Individual pretext objectives have complementary limitations: generative models emphasize local reconstruction fidelity but may underemphasize global semantic invariance, and contrastive models learn strong global representations but may lack dense spatial precision. Hybrid learning jointly optimizes multiple objectives within a unified training scheme.

\subsubsection{Contrastive and generative integration}

DiRA jointly optimizes contrastive alignment, pixel-wise reconstruction, and a \gls{gan}-style discriminator to enforce anatomical realism, capturing both global structure and local detail~\cite{haghighi2024self}. MAEMC-NET pairs \gls{mim} with MoCo-style contrastive learning for pulmonary nodule analysis: a \gls{vit}-Large reconstructs masked patches in axial \gls{ct} slices while simultaneously aligning axial and coronal composite patches from the same lesion, giving both intra-slice continuity and cross-plane discrimination~\cite{zhao2025maemc}. Khalid et al.\ combine pixel-wise reconstruction with ``clean vs.\ perturbed'' and ``original vs.\ rotated'' classification tasks for coronary artery disease diagnosis in cardiac \gls{mri}, preserving anatomical appearance while maintaining sensitivity to orientation and noise~\cite{khalid2025improving}. SSMTL uses adjacent-slice structural similarity: it reconstructs two neighboring slices at a fixed distance from an anchor slice while a Siamese branch maximizes cosine similarity between two slices from the same scan~\cite{dong2021self}. PCRLv2 is built explicitly around visual-information preservation: it augments siamese feature comparison with multi-scale pixel restoration on a feature pyramid, encoding pixel-level, semantic, and scale information together, and reports large gains on brain tumor segmentation, chest pathology identification, pulmonary nodule detection, and abdominal organ segmentation under limited annotation~\cite{zhou2023pcrlv2}. Whereas PCRLv2 realizes information preservation within a single framework, the present review adopts it as a cross-study analytical lens (Section~\ref{sec:introduction}).

\subsubsection{Coupling contrastive and predictive objectives}

Li et al.\ jointly optimize rotation prediction and multi-view contrastive learning for retinal disease diagnosis, decoupling the feature space into rotation-related structural components (vessels, optic disc) and rotation-invariant components~\cite{li2021rotation}. Wang et al.\ couple SimSiam-style~\cite{chen2021simsiam} representation matching with a masked-jigsaw predictive objective for histopathology segmentation, processing paired low-resolution context and high-resolution target patches through separate branches, with a Context--Target Fusion module masking and shuffling target features before concatenation~\cite{wang2024multi}.

\subsubsection{Sequential and nested task learning}

Some frameworks apply pretext tasks sequentially. Nested \gls{ssl} for white matter tract segmentation first pretrains an encoder using voxel-wise streamline density regression from automated tractography, then initializes a second pretext task that predicts 72 atlas-propagated tract masks, refining tract-specific contextual anatomy before fine-tuning~\cite{lu2021volumetric}.

\subsubsection{Graph-based and multi-task hybrid frameworks}

Several frameworks combine three or more objectives. Song et al.\ combine rotation prediction, multi-view contrastive learning, and reconstruction for COVID-19 lesion segmentation~\cite{song2022covid}, and the two-stage transformer of Qayyum et al.\ uses masked volume inpainting, contrastive learning on augmented views of the same 3D patch, and rotation prediction as proxy tasks in a self-supervised first stage, before supervised fine-tuning of a multiview multi-scale parallel-attention U-Net for brain-tumor, head-and-neck, and cardiac segmentation~\cite{qayyum2023two}. The earlier surrogate supervision framework combines 3D flip/rotation prediction, Wasserstein \gls{gan}-based patch reconstruction, and conditional \gls{gan} colorization~\cite{tajbakhsh2019surrogate}. Pathology-specific multi-tasking is exemplified by Self-Path, which integrates multi-scale magnification prediction and magnification-aware jigsaw with agnostic tasks (rotation, flipping, autoencoding, gradient-reversal domain prediction) to learn scale-aware morphology~\cite{koohbanani2021self}. Retinal fuzzy clustering combines autoencoder reconstruction, fuzzy C-means, and an \gls{ssl} membership regression head~\cite{luo2020retinal}, and ColorMe combines pixel-wise green-red/blue colorization with red/blue color distribution estimation via Kullback--Leibler divergence for endoscopy~\cite{li2020multi}.

Two more recent graph- and modality-oriented frameworks are worth calling out. Wang et al.\ apply hybrid graph-level \gls{ssl} to functional \gls{mri} by combining denoising diffusion on functional connectivity graphs with SimSiam-style~\cite{chen2021simsiam} contrastive alignment between original and diffusion-augmented graph embeddings, learning disease-relevant brain connectivity without manual labels~\cite{wang2025self}. UniMiSS+ handles unpaired 2D X-rays and 3D \gls{ct} volumes through a dimension-free pyramid Transformer trained with DINO-style~\cite{caron2021dino} view consistency, slice--volume consistency, \gls{ct}--\gls{drr} cross-modality matching, and reconstruction~\cite{xie2024unimiss+}.

A field-defining large-scale example is Swin UNETR, whose 3D Swin-Transformer encoder is pretrained on 5{,}050 \gls{ct} volumes with three simultaneous proxy tasks: masked volume inpainting, contrastive learning, and rotation prediction, capturing texture, discriminative, and structural cues respectively. Its scale and multi-objective design make it a standard volumetric \gls{ssl} baseline~\cite{tang2022self}.

\paragraph{Synthesis.} Hybrid methods combine complementary signals to preserve both
global discriminability and local structure, and are the most consistent performers across tasks. Large-scale volumetric frameworks (PCRLv2, Swin UNETR) show the approach scales. The costs are training complexity, more hyperparameters, and difficulty attributing gains to any single objective. Evidence is strong for classification and segmentation but sparse for reconstruction and regression. The real importance of these categories depends on how their representations transfer to clinical tasks, which the next section analyzes systematically.

\section{Task alignment analysis}
\label{sec:task-alignment-analysis}

This section analyzes the reviewed studies from a task-alignment perspective. We first summarize the distribution of downstream tasks, then examine how \gls{ssl} objective families align with classification, segmentation, detection, reconstruction, and regression, and finally discuss how modality, label availability, and transfer behavior affect \gls{ssl} effectiveness.

\subsection{Distribution of downstream tasks}

Table~\ref{tab:distribution} reports the distribution of the reviewed studies across the four \gls{ssl} objective families. Contrastive learning is the most frequent (22 studies), while the remaining three families are represented by 20, 19, and 17 studies. The relatively even spread indicates that the field is no longer dominated by a single paradigm.

\begin{table}[ht]
    \centering
    \caption{Distribution of the $78$ reviewed studies across \gls{ssl} objective families.}
    \label{tab:distribution}
    \small
    \begin{tabular}{@{}lc@{}}
    \toprule
    \textbf{Objective family} & \textbf{No. of studies} \\
    \midrule
    Contrastive learning                       & 22 \\
    Non-contrastive \& predictive learning      & 20 \\
    Generative \& reconstruction-based learning & 19 \\
    Hybrid learning                             & 17 \\
    \midrule
    \textbf{Total}                              & \textbf{78} \\
    \bottomrule
    \end{tabular}
\end{table}

Classification is the most frequently studied downstream task, spanning dermatology, chest X-ray, \gls{ct}, \gls{mri}, fundus, histopathology, ultrasound, dental radiography, and functional \gls{mri}, with applications including skin disease recognition, pulmonary nodule malignancy prediction, COVID-19 diagnosis, prostate cancer diagnosis, glaucoma detection, HER2 grading, and brain disorder detection~\cite{azizi2021big, ali2023self, zhao2025maemc, de2025self, sriram2021covid, chen2021momentum, gryshchuk2025contrastive, yi2024parcl, nguyen2023self, jamaludin2017self, li2020self, mohammad2024artificial, hirsch2023self, nasiri2024vim4path, taleb2022self, dong2023swssl, daydar2024med, manna2023self, xing2023self, wang2023pyramid, wolf2023self, ozbay2024kidney, hervella2021self, hervella2022multimodal, chyrmang2025self, khalid2025improving, li2021rotation, song2022covid, tajbakhsh2019surrogate, koohbanani2021self, luo2020retinal, wang2025self, xie2024unimiss+, zhou2023pcrlv2}. Segmentation is the second major category, particularly for volumetric and structurally complex data (brain tumor, multi-organ \gls{ct}, cardiac \gls{mri}, retinal vessels, stroke lesions, dental structures, histopathology nuclei, white matter tracts)~\cite{zhu2020rubik, meng2024self, goncharov2023vox2vec, ma2025multi, liu20243d, hu2025self, xie2020pgl, ye2024cads, zhuang2019self, tao2020revisiting, nguyen2020self, bai2019self, spitzer2018improving, moreau2025self, taleb2021multimodal, almalki2023self, almalki2024self, zhou2019models, chen2019self, dominic2023improving, tahghighi2024enhancing, hervella2020learning, morano2020multimodal, ross2018exploiting, haghighi2024self, qayyum2023two, dong2021self, wang2024multi, lu2021volumetric, song2022covid, tajbakhsh2019surrogate, li2020multi, xie2024unimiss+, wu2024voco, zhou2023pcrlv2, tang2022self}. Because segmentation requires dense spatial prediction and accurate boundary preservation, it is a more informative testbed for representation quality than classification.

A smaller subset of studies targets detection and anomaly localization~\cite{mh2023lvm, jiao2020self, almalki2023self, patel2023self, haghighi2024self, sriram2021covid, zhou2023pcrlv2}, reconstruction (dominated by undersampled \gls{mri})~\cite{ekanayake2025cl}, regression (body-part position, visual acuity, biomarkers, cellularity, perfusion, survival)~\cite{ciga2022self, suekei2024multi, zhang2017self, taleb2021multimodal, cho2023chess, liu2025perfusion}, and localization, saliency, forecasting, and modality synthesis~\cite{suekei2024multi, chen2019self, hervella2020learning, cao2020auto}. Several studies evaluate representations across multiple downstream tasks jointly~\cite{zhu2020rubik, ciga2022self, cho2023chess, mh2023lvm, zhuang2019self, nguyen2020self, taleb2021multimodal, zhou2019models, chen2019self, haghighi2024self, song2022covid, tajbakhsh2019surrogate, li2020multi, xie2024unimiss+}, which provides useful evidence about whether learned features are narrow or broadly transferable.

Two takeaways follow. First, medical \gls{ssl} remains primarily bench-marked on classification and segmentation. Second, the growing presence of multi-task evaluation shows that medical \gls{ssl} must support a diverse set of representation requirements, which motivates the alignment analysis below.

\subsection{Alignment between SSL objectives and downstream tasks}

Table~\ref{tab:alignment} summarizes the qualitative alignment between \gls{ssl} objectives and major downstream tasks. Ratings synthesize recurring trends across the reviewed studies rather than absolute performance scores. Each cell reflects the direction, strength, and consistency of transfer outcomes for the corresponding family--objective pairing. \ding{108} indicates strong and well-supported alignment, based on multiple studies showing consistent positive transfer and a clear match between the encouraged representational property and the downstream demand. \ding{109} indicates moderate or context-dependent alignment, where evidence is positive but limited, mixed, contingent on modality or label regime, or supported by only one study. \ding{55} indicates weak alignment or a risk of negative transfer, where no benefit or measurable degradation is reported, or the objective is expected to suppress required information. \texttt{--} indicates that the pairing was not evaluated in the reviewed studies. Because datasets, architectures, and evaluation protocols vary substantially across studies, these ratings should be read as a balance of evidence, not a quantitative synthesis.

The subsections below do not re-describe the methods of Section~\ref{sec:taxonomy-of-pretext-tasks-in-medical-imaging}. Instead, each states the direction of transfer for a family--objective pairing, the condition under which it holds, and the strength of the supporting evidence.

\begin{table*}[tb]
    \centering
    \caption{Task-alignment matrix mapping \gls{ssl} objective families to downstream clinical objectives. \ding{108}~strong and well-supported, \ding{109}~moderate or context-dependent, \ding{55}~weak or prone to negative transfer, \texttt{--}~not evaluated for this pairing in the reviewed studies. Symbols marked $\dagger$ reflect the authors' interpretive judgment where direct study-level evidence is limited; all other ratings summarize the direction of transfer reported across multiple reviewed studies. The supporting evidence for each rating is given in the text below.}
    \label{tab:alignment}
    \small
    \resizebox{\textwidth}{!}{%
    \begin{tabular}{@{}lccccc@{}}
    \toprule
    \textbf{Objective family} & \textbf{Classification} & \textbf{Segmentation} 
    & \textbf{Detection /} & \textbf{Reconstruction} & \textbf{Regression} \\
     & & & \textbf{anomaly} & & \\
    \midrule
   Contrastive (instance / patient-level) & \ding{108} & \ding{109} & \ding{109} & \ding{55}$\dagger$ & \ding{109} \\
    Contrastive (local / voxel-level)      & \ding{109} & \ding{108} & \texttt{--} & \texttt{--} & \texttt{--} \\
    Non-contrastive \& predictive          & \ding{109} & \ding{108} & \ding{109} & \texttt{--} & \ding{109} \\
    Generative / reconstruction            & \ding{109} & \ding{108} & \ding{108} & \ding{108}$\dagger$ & \ding{109} \\
    Hybrid                                 & \ding{108} & \ding{108} & \ding{109} & \texttt{--} & \texttt{--} \\
    \bottomrule
    \end{tabular}%
    }
    \par\vspace{2pt}
    \footnotesize\raggedright
    Note: Cross-modal/structural and acquisition-based contrastive variants (Sections~\ref{subsubsec:crossmodal}--\ref{subsubsec:kspace}) are modality- or task-specific and rest on comparatively few studies. They are discussed individually in the text rather than rated here; the acquisition-based variant aligns specifically with reconstruction.
\end{table*}

\subsubsection{Contrastive learning and classification}

Instance- and patient-level contrastive methods transfer positively to classification, but the mechanism is conditional: they induce globally discriminative representations \emph{when} positive-pair construction and augmentation preserve target-relevant semantics. Where they do, transfer is strong and holds into extreme few-shot regimes---86\% liver-view accuracy from one labeled image per class~\cite{ali2023self}, and domain-specific prostate \gls{mri} pretraining exceeding ImageNet baselines~\cite{de2025self}, with patient-aware variants adding further gains~\cite{azizi2021big, nguyen2023self}. Where augmentation or pair construction discards the diagnostic signal (Section~\ref{subsubsec:instance}), the same objective underperforms. Evidence here is the strongest and most consistent in the review.

\subsubsection{Spatial prediction and segmentation}

Spatial and anatomical prediction transfers positively to segmentation, because the objective is forced to encode the structural and boundary relations that dense prediction requires~\cite{zhu2020rubik, zhuang2019self, bai2019self, nguyen2020self}. The condition is specificity: the more the pretext target reflects the anatomy of the downstream task, the stronger the transfer, so atlas- and position-guided tasks outperform generic geometric ones---ASPECTS yields stable low-contrast stroke-lesion segmentation~\cite{moreau2025self} and geodesic-distance prediction aligns with cytoarchitectonic parcellation~\cite{spitzer2018improving}. As noted in Section~\ref{subsubsec:spatial}, part of this strength reflects weak anatomical supervision rather than self-supervision alone. Evidence is strong for segmentation and weaker for classification.

\subsubsection{Generative and reconstruction tasks for dense prediction}

Reconstruction-based objectives transfer positively to dense prediction, detection, and reconstruction, because their pixel-level targets retain the fine-grained local detail these tasks need~\cite{xing2023self, wang2023pyramid, zhou2019models, chen2019self, dominic2023improving}. The alignment extends to anomaly detection, where modeling the normal-data distribution supports pathological-uptake detection in PET~\cite{patel2023self}. Direct evidence for reconstruction is limited to a contrastive acquisition-based k-space objective rather than a generative one~\cite{ekanayake2025cl}. Accordingly, the strong generative--reconstruction rating in Table~\ref{tab:alignment} reflects an interpretive judgment (Section~\ref{subsubsec:mim}). In dental imaging, masked image modeling aligns with both teeth numbering and restoration detection~\cite{almalki2023self}. The condition is that corruption must not destroy the diagnostic signal (Section~\ref{subsubsec:mim}). Evidence is strong for segmentation and reconstruction and growing for detection.

\subsubsection{Cross-modal learning for complex clinical tasks}

Cross-modal objectives transfer positively when the paired modalities carry complementary, disease-relevant information. In retina, synthesizing fluorescein angiography from color retinography improves vessel segmentation and diabetic retinopathy grading by encouraging representations predictive of vascular structures shared across the paired modalities~\cite{hervella2020learning, morano2020multimodal, hervella2021self, hervella2022multimodal}, and in volumetric imaging, aligning 3D \gls{ct} with synthesized 2D \glspl{drr} improves 3D kidney and liver segmentation while reducing pretraining time~\cite{ye2024cads}. The condition is that the cross-modal relationship be modeled explicitly: naive modality concatenation degrades transfer (Section~\ref{subsubsec:negative}). Evidence rests on comparatively few but consistent studies.

\subsubsection{Hybrid approaches for generalizable representations}

Hybrid objectives transfer positively across multiple downstream tasks, because combining complementary signals covers demands no single objective meets: DiRA gives large gains in 1\% label regimes for X-ray and \gls{ct}~\cite{haghighi2024self}, MAEMC-NET pairs reconstruction with contrast for pulmonary-nodule analysis~\cite{zhao2025maemc}, SSMTL improves 3D cardiac and brain-tumor segmentation~\cite{dong2021self}, and pathology-specific multi-task frameworks show superior domain adaptation~\cite{tajbakhsh2019surrogate, koohbanani2021self}. The condition is complementarity: gains appear only when the combined objectives cover genuinely different requirements, and large-scale volumetric frameworks (PCRLv2~\cite{zhou2023pcrlv2}, Swin UNETR~\cite{tang2022self}) show the pattern scales. Evidence is strong for classification and segmentation but sparse for reconstruction and regression.

\subsection{Effect of imaging modality}

The same \gls{ssl} objective does not perform equally well across modalities, because the type of clinically relevant information differs.

Volumetric modalities (\gls{ct}, \gls{mri}) benefit from pretext tasks that preserve spatial structure and anatomical consistency, particularly for segmentation~\cite{zhu2020rubik, zhuang2019self, tao2020revisiting, nguyen2020self, bai2019self}. For \gls{ct} specifically, the appropriate masking ratio is target-dependent rather than uniformly low: aggressive masking degrades detection of small pulmonary nodules~\cite{zhao2025maemc}, whereas a large controlled study reported high static and dynamic ratios (up to 90\%) to be equally effective on volumetric data~\cite{wald2025revisiting}, so the optimal ratio tracks the spatial scale of the target pathology (Section~\ref{subsubsec:mim}). \gls{mri} additionally benefits from methods that handle multi-sequence or multi-view relationships, especially in brain, cardiac, prostate, and musculoskeletal imaging~\cite{bai2019self, taleb2021multimodal}, and reconstruction benefits from acquisition-aware objectives applied directly in k-space~\cite{ekanayake2025cl}. At scale, large-scale contrastive (VoCo~\cite{wu2024voco}), multi-task (Swin UNETR~\cite{tang2022self}), and self-distillation (3DINO~\cite{xu2025_3dino}) frameworks confirm that spatially aware, 3D-consistent objectives are the dominant choice for volumetric data across $10^4$--$10^5$ volumes.

For projection-based X-ray imaging, the global anatomical layout is stable but pathological features may occupy small regions. Contrastive, redundancy-reduction, and \gls{mim} methods all transfer provided they preserve fine-grained diagnostic structures. Among these, the masked-modeling studies that examine masking ratio directly report better retention of small findings at reduced ratios in dental and chest radiography~\cite{xing2023self, almalki2023self}.

Histopathology is dominated by contrastive learning, self-distillation, multi-resolution learning, and pathology-specific hybrid pretext tasks, because local texture, stain variation, and multi-resolution tissue patterns matter more than volumetric continuity~\cite{ciga2022self, yan2022deep, nasiri2024vim4path, wang2023pyramid, chyrmang2025self, wang2024multi, koohbanani2021self}. Pathology-specific tasks such as JigMag show especially strong gains in low-label and domain-adaptation settings~\cite{koohbanani2021self}. This modality also anchors most foundation-scale pretraining in medical imaging: contrastive (CTransPath~\cite{wang2022ctranspath}) and self-distillation models (HIPT~\cite{chen2022hipt}, Virchow~\cite{vorontsov2023virchow}, Prov-GigaPath~\cite{xu2024provgigapath}) pretrained on up to $10^9$ patches confirm that patch-level and multi-resolution objectives dominate this modality.

Retinal imaging has an unusually strong role for cross-modal \gls{ssl} because clinically meaningful information is often distributed across fundus, fluorescein angiography, and \gls{oct}~\cite{yi2024parcl, nguyen2023self, li2020self, suekei2024multi, hervella2020learning, morano2020multimodal, hervella2021self, hervella2022multimodal, li2021rotation, luo2020retinal}. Cross-modal reconstruction or alignment improves glaucoma detection, diabetic retinopathy grading, vessel segmentation, and prognosis prediction. At scale, RETFound learns generalizable representations from 1.6 million unlabeled retinal images via masked autoencoding and improves label efficiency across ocular and systemic disease detection~\cite{zhou2023retfound}.

Ultrasound and endoscopic video contain temporal dynamics, operator dependence, and motion variation, and benefit most from temporal ordering, frame consistency, or video-based representation learning~\cite{ali2023self, vanberlo2023exploring, hirsch2023self, jiao2020self, ross2018exploiting}. Frame ordering and phase recognition are especially effective in low-label settings.

Because foundation-scale models span contrastive, generative, and self-distillation objectives, this evidence corroborates rather than displaces the family-level patterns above. These models (CTransPath, HIPT, Virchow, Prov-GigaPath, RETFound, 3DINO) are cited as supporting context and are not part of the $78$ study count in Table~\ref{tab:distribution}. VoCo, PCRLv2, and Swin UNETR, which are analyzed at the study level in Section~\ref{sec:taxonomy-of-pretext-tasks-in-medical-imaging}, are included in that count. Therefore, task alignment in medical \gls{ssl} has both a task and a modality axis: an objective that transfers well within one modality may not transfer to another with different structural priors or clinically relevant features.

\subsection{Influence of label regime}

\gls{ssl} provides its largest gains when labeled data are scarce. In extreme low-label settings, contrastive learning gives strong results with a handful of labeled examples: SimCLR on ultrasound liver-view classification reaches 86\% accuracy with one labeled image per class and over 98\% with small labeled subsets~\cite{ali2023self}, and contrastive few-shot COVID-19 diagnosis on \gls{ct} performs well in one- and five-shot regimes~\cite{chen2021momentum}. Similar patterns appear in segmentation. Rubik's Cube-based \gls{ssl} trained with 50\% labels matches models trained from scratch on 100\%~\cite{zhu2020rubik}, and cardiac \gls{mri} anatomical position prediction gives substantial Dice improvements with few annotated subjects~\cite{bai2019self, tao2020revisiting, chen2019self}. For necrotic lung lesion segmentation, contrastive learning with 10\% labeled data surpasses fully supervised training on the full dataset~\cite{liu20243d}; context-restoration \gls{ssl} matches full supervision on brain tumor segmentation with 25--50\% labels~\cite{chen2019self}. Nested \gls{ssl} for diffusion \gls{mri} tract segmentation performs well with as few as 2--5 labeled scans~\cite{lu2021volumetric}, and multi-modal \gls{ssl} for retinal vessel segmentation is effective with one or two labeled images~\cite{hervella2020learning, morano2020multimodal}.

The gains attenuate but do not vanish as labels increase. Large-scale histopathology \gls{ssl} outperforms ImageNet pretraining even with full labeled datasets~\cite{ciga2022self}, and \gls{mae}-based and hybrid \gls{ssl} frameworks improve classification and segmentation under full supervision~\cite{xing2023self, haghighi2024self, xie2024unimiss+}, though by smaller margins. A small number of studies use \gls{ssl} representations directly without downstream supervision: contrastive clustering in histopathology~\cite{yan2022deep} and fuzzy clustering in retinal imaging~\cite{luo2020retinal} both achieve strong performance in fully unsupervised settings. Overall, \gls{ssl} is most valuable when unlabeled data are abundant and labels are scarce, which is the typical situation in medical imaging.

\subsection{Positive and negative transfer}

\subsubsection{Positive transfer}

Most reviewed studies report positive transfer, in which \gls{ssl}-pretrained models outperform training from scratch or ImageNet initialization. These outcomes are the results reported by the original authors and were not independently reproduced; because studies reporting successful transfer are more likely to be published, this pattern should be read as the prevailing direction of available evidence rather than an unbiased estimate.

Representative gains include: MICLe improving dermatology classification by 6.7\%~\cite{azizi2021big}; SimCLR-based ultrasound classification reaching 98.7\% accuracy~\cite{ali2023self}; Rubik's Cube prediction improving classification accuracy by 8.11\% and segmentation Dice by more than 8\%~\cite{zhu2020rubik}; SimCLR histopathology features outperforming ImageNet with average F1 improvements exceeding 28\% in linear evaluation~\cite{ciga2022self}; and Models Genesis and UniMiSS+ transferring consistently across \gls{ct}, \gls{mri}, and X-ray~\cite{zhou2019models, xie2024unimiss+}. In extreme low-label regimes, nested \gls{ssl} enables accurate white matter tract segmentation with 2--5 labeled scans~\cite{lu2021volumetric}, and multi-modal retinal \gls{ssl} performs well with one or two labeled images~\cite{hervella2020learning, morano2020multimodal}.

\subsubsection{Negative or limited transfer}
\label{subsubsec:negative}

Several studies document situations where transfer is limited or actively harmful, and the recurring causes are the two failure modes identified in the taxonomy: shortcut learning and augmentation- or corruption-induced loss of diagnostic signal.

The first is \emph{shortcut learning}. Because the pretraining objective is unconstrained by labels, an encoder can satisfy it using acquisition-correlated cues, including scanner, institution, or protocol signatures, rather than anatomy, which inflates in-distribution performance while degrading cross-site transfer (Section~\ref{subsubsec:instance}; see also Section~\ref{sec:open-challenges-and-limitations}). A related instance-level effect is the false-negative problem, where treating scans of the same organ, anatomy, or patient as negatives pushes biologically similar cases apart (Section~\ref{subsubsec:instance}), and the bilateral and longitudinal patient-level pairings that may enforce similarity between a diseased and a healthy view, or across time points that straddle disease onset (Section~\ref{subsubsec:patient}). Consistent with this, naively concatenating modalities in contrastive multi-modal retinal analysis reduced performance by 5--6\%, indicating that unmodeled modality relationships hurt representation learning~\cite{li2020self}.

The second is \emph{loss of diagnostic signal through augmentation or corruption}. \gls{mim} in pulmonary \gls{ct} and dental radiography degrades when the masking ratio is too high, because small, localized pathological features are easily removed under aggressive masking (Section~\ref{subsubsec:mim})~\cite{zhao2025maemc, almalki2023self}, and augmentation policies that crop or jitter too aggressively can suppress micro-calcifications, small tumors, or fine vascular structures (Section~\ref{subsubsec:instance})~\cite{yi2024parcl, taleb2022self}. A distinct alignment failure appears in voxel-level contrastive \gls{ct} segmentation, where features learned from structural anatomy transferred well to organs but weakly to some tumor segmentation tasks without fine-tuning~\cite{goncharov2023vox2vec}, showing that anatomically-driven features do not automatically capture pathological variation.

Finally, when large labeled datasets are available, the \gls{ssl}--supervised gap narrows because task-specific representations can be learned directly from labels~\cite{ciga2022self, cho2023chess}.

\section{Practical design guidelines for task-aligned SSL in medical imaging}
\label{sec:practical-design-guidelines}

The reviewed evidence indicates that no single \gls{ssl} strategy dominates across medical imaging applications. Effective design is task-aware, pathology-aware, and modality-aware rather than directly adapted from general computer vision. Table~\ref{tab:guidelines} summarizes practical recommendations linking each downstream goal to a recommended class of \gls{ssl} objective. The subsections below give the corresponding guidance for modality, label regime, hybrid design, and negative-transfer avoidance.

\begin{table*}[tb]
    \centering
    \caption{Practical suggestions for selecting \gls{ssl} objectives according to the downstream clinical task.}
    \label{tab:guidelines}
    \small
    \resizebox{\textwidth}{!}{%
    \begin{tabular}{@{}p{2.5cm}p{3.9cm}p{4.0cm}p{3.4cm}@{}}
    \toprule
    \textbf{Downstream goal} & \textbf{Suggested \gls{ssl} objectives} & \textbf{Rationale} 
    & \textbf{Representative studies} \\
    \midrule
    Classification & Instance- or patient-level contrastive learning 
    & Promotes global semantic discriminability and class separability 
    & Azizi et al.~\cite{azizi2021big}, Ali et al.~\cite{ali2023self}, de Almeida et al.~\cite{de2025self} \\[4pt]
    Segmentation & Spatial / anatomical prediction; local or voxel-level contrastive 
    & Preserves anatomical boundaries and inter-region spatial structure 
    & Bai et al.~\cite{bai2019self}, Goncharov et al.~\cite{goncharov2023vox2vec}, Tao et al.~\cite{tao2020revisiting} \\[4pt]
    Detection \& anomaly localization & Masked image modeling; context restoration 
    & Retains fine-grained local texture and contextual cues 
    & Almalki \& Latecki~\cite{almalki2023self}, Patel et al.~\cite{patel2023self}, Zhou et al.~\cite{zhou2019models} \\[4pt]
    Video / temporal analysis & Temporal-order or frame-consistency prediction 
    & Captures motion and inter-frame anatomical transitions 
    & Jiao et al.~\cite{jiao2020self}, Hirsch et al.~\cite{hirsch2023self} \\[4pt]
    Multimodal diagnosis & Cross-modal alignment or synthesis 
    & Learns modality-invariant yet complementary representations 
    & Suekei et al.~\cite{suekei2024multi}, Hervella et al.~\cite{hervella2021self}, Taleb et al.~\cite{taleb2021multimodal} \\
    \bottomrule
    \end{tabular}%
    }
\end{table*}

\subsection{Match the SSL objective to the downstream task}

The recommendations below are conditional rather than universal: each names a plausible candidate objective together with the condition under which it is appropriate.

For classification, instance- or patient-level contrastive learning is a reasonable candidate when classification depends primarily on augmentation-stable semantics and positive pairs do not erase localized pathology~\cite{azizi2021big, ali2023self, de2025self, sriram2021covid, chen2021momentum, gryshchuk2025contrastive, yi2024parcl, nguyen2023self}. Segmentation generally requires both local boundary fidelity and global anatomical context. Local contrastive, anatomical-position, and multi-scale objectives are therefore plausible candidates, with the appropriate balance depending on whether the target is an organ, vessel, or lesion~\cite{zhu2020rubik, tao2020revisiting, nguyen2020self, bai2019self, goncharov2023vox2vec, ma2025multi}. Detection and anomaly localization depend on subtle local patterns, so masked image modeling and context restoration are suitable when corruption preserves the diagnostic signal~\cite{dong2023swssl, almalki2023self, zhou2019models, chen2019self}. Video and sequential imaging benefit from temporal-order or frame-consistency objectives~\cite{hirsch2023self, jiao2020self, manna2023self}, and multi-modal diagnosis benefits from cross-modal alignment or synthesis when the paired modalities carry complementary information~\cite{suekei2024multi, taleb2021multimodal, hervella2020learning, morano2020multimodal, hervella2021self, hervella2022multimodal}. The overarching rule is that the pretext task should preserve, not suppress, the information the downstream objective needs.

\subsection{Incorporate modality-specific priors}

Medical images differ from natural images in dimensionality, noise, structural priors, and acquisition physics, and \gls{ssl} designs that reflect these differences transfer better.

For \gls{ct} and \gls{mri}, use volumetric and anatomical priors---slice ordering, anatomical position prediction, voxel-level contrastive learning, and spatial puzzle solving---which support both segmentation and structural classification~\cite{zhu2020rubik, xie2020pgl, tao2020revisiting, nguyen2020self, bai2019self, goncharov2023vox2vec}. For histopathology, use patch-based contrastive learning and multi-resolution objectives that capture morphology across scales~\cite{ciga2022self, nasiri2024vim4path, wang2023pyramid, wang2024multi}. For retinal imaging, favor pathology-aware or multi-modal pretext tasks that focus on clinically meaningful regions and cross-modality anatomy~\cite{yi2024parcl, li2020self, suekei2024multi, hervella2020learning, hervella2021self, hervella2022multimodal}. For X-ray, tune masking and augmentation against modality-specific redundancy and pathology scale rather than adopting natural-image defaults. Masked-modeling studies report loss of small features under aggressive masking~\cite{xing2023self, almalki2023self}. For ultrasound and endoscopic video, use temporal or sequence-based objectives such as frame ordering and temporal prediction, which model motion and anatomical transitions~\cite{hirsch2023self, jiao2020self, manna2023self}.

\subsection{Match the strategy to the label regime}

\gls{ssl} gives the largest gains when labeled data are scarce and unlabeled data are abundant~\cite{ali2023self, chen2021momentum, bai2019self, lu2021volumetric, tajbakhsh2019surrogate}. In few-shot or small-sample settings, contrastive and reconstruction-based methods let models learn anatomy and structure before fine-tuning, often matching or exceeding fully supervised baselines with a small fraction of the labels~\cite{ali2023self, chen2021momentum, bai2019self, hervella2020learning, morano2020multimodal}. In moderately labeled scenarios, \gls{ssl} initialization outperforms ImageNet and random initialization and speeds up convergence, particularly for 3D data whose structural properties diverge sharply from natural images~\cite{azizi2021big, de2025self, cho2023chess, mh2023lvm, zhou2019models, tajbakhsh2019surrogate}. Even with abundant labels, \gls{ssl} improves cross-dataset transfer, distribution-shift robustness, and generalization across institutions~\cite{azizi2021big, sriram2021covid, suekei2024multi, mh2023lvm, xie2024unimiss+}. Because clinical archives typically contain large volumes of unlabeled imaging, \gls{ssl} is a natural strategy for extracting value from these resources~\cite{ciga2022self, mh2023lvm, zhou2019models}.

\subsection{Combine objectives when signals are complementary}

Hybrid and multi-modal frameworks may produce broader representations than single-objective methods when the component objectives encode genuinely complementary information and are validated through matched ablations~\cite{zhao2025maemc, haghighi2024self, qayyum2023two, dong2021self}. Pairing contrastive and generative objectives lets the model capture both semantic separation and dense structural detail, which improves both classification and segmentation across \gls{ct}, \gls{mri}, and chest X-ray~\cite{zhao2025maemc, haghighi2024self, qayyum2023two}. Multi-modal pretraining is particularly valuable when modalities carry complementary clinical information---fundus with fluorescein angiography or \gls{oct}~\cite{hervella2020learning, morano2020multimodal, hervella2021self, hervella2022multimodal}, or \gls{ct} with X-ray or across \gls{mri} sequences~\cite{suekei2024multi, taleb2021multimodal}. Cross-dimensional pretraining that jointly uses 2D slices and 3D volumes is a related direction, enabling knowledge transfer across imaging formats~\cite{mh2023lvm, xie2024unimiss+}. Sequential or nested multi-task pipelines can further improve label efficiency and robustness to domain shift~\cite{dong2021self, lu2021volumetric}.

\subsection{Design SSL objectives to minimize negative transfer}

Poorly aligned pretext tasks can produce limited or negative transfer, so the specific failure modes discussed in Section~\ref{sec:task-alignment-analysis} deserve explicit design attention. Augmentation policies should preserve subtle pathological patterns rather than adopting aggressive natural-image defaults, which can suppress micro-calcifications, small tumors, or fine vascular structures~\cite{yi2024parcl, li2020self, taleb2022self}. Masking ratios for medical \gls{mim} should be matched to the spatial scale of the diagnostic target: aggressive masking can destroy small, sparse findings, though high ratios remain effective where anatomy is densely distributed and the target is large~\cite{xing2023self, almalki2023self, wald2025revisiting}. Pretext objectives should also match the spatial or temporal structure that the downstream task requires: global instance discrimination is insufficient for segmentation~\cite{goncharov2023vox2vec, zhuang2019self, nguyen2020self, bai2019self}, and static objectives are insufficient for video~\cite{jiao2020self, taleb2021multimodal}. Evaluating pretrained representations across multiple downstream tasks and datasets is a useful check that learned features generalize rather than overfit to the pretraining objective.

\section{Open challenges and limitations}
\label{sec:open-challenges-and-limitations}

Despite rapid progress, several challenges limit the clinical adoption of \gls{ssl} in medical imaging.

\textbf{Generalizability across medical distributions.} Most \gls{ssl} methods are evaluated on data from a single institution or acquisition protocol, largely because medical data are not publicly shareable at scale. Medical images are highly sensitive to scanner type, imaging parameters, patient population, and clinical workflow, so models trained on one distribution can degrade elsewhere. Although \gls{ssl}-pretrained models generally report greater robustness than fully supervised counterparts, systematic multi-institution evaluation remains rare, and consistent cross-institution performance is an open requirement for real-world deployment. A related risk is shortcut learning: because \gls{ssl} objectives are unconstrained by labels, they can latch onto scanner-, protocol-, or institution-specific signatures rather than anatomy, which inflates in-distribution performance while worsening cross-site transfer.

\textbf{Isolating the contribution of \gls{ssl}.} Recent \gls{ssl} studies often introduce more powerful architectures, larger backbones, or improved fine-tuning strategies alongside the \gls{ssl} objective. When multiple factors change simultaneously, it is difficult to attribute performance gains to the \gls{ssl} objective itself rather than to architectural capacity. Standardized evaluation protocols, with common backbones and baselines, are needed to measure the independent contribution of the pretext design.

\textbf{Computational cost.} Many \gls{ssl} approaches depend on large batch sizes, long training schedules, and high-capacity architectures, and require multiple \glspl{gpu} that are not routinely available in clinical research environments. Medical images amplify this cost because histopathology is high-resolution and \gls{ct} and \gls{mri} are volumetric, both of which increase memory and compute requirements. More efficient training strategies are needed to broaden access.

\textbf{Task alignment and label-regime dependence.} As Sections~\ref{sec:task-alignment-analysis}--\ref{sec:practical-design-guidelines} discuss in detail, misaligned pretext tasks can produce negative transfer, and the magnitude of \gls{ssl} gains depends strongly on the label regime. Both effects complicate the question of when \gls{ssl} is worth adopting for a given clinical problem and reinforce the need for systematic evaluation across label availability scenarios.

\textbf{Scope and limitations of this review.} As a narrative, task-oriented synthesis, this review prioritizes conceptual coverage of how \gls{ssl} objective design relates to downstream clinical objectives over exhaustive enumeration. The 78 studies were selected to be representative of major paradigms, modalities, and downstream tasks, so the emphasis reflects the areas where task-alignment effects are most clearly documented, and some related work falls outside the scope. Because the reviewed studies differ substantially in datasets, architectures, backbones, and evaluation metrics, a quantitative meta-analysis was neither feasible nor appropriate; the ratings in Table~\ref{tab:alignment} therefore summarize the prevailing direction of evidence and are intended as design guidance rather than benchmark comparisons. Our observations are based on results as reported by the original authors, which further motivates the standardized evaluation protocols recommended above.

\textbf{Evaluation integrity: leakage, contamination, and benchmark saturation.} Because \gls{ssl} pretrains on large unlabeled corpora, it is especially exposed to data leakage. When slices from the same patient, study, or institution appear in both pretraining and evaluation, or when public pretraining corpora overlap with downstream test sets, reported transfer is optimistically biased. Repeated reuse of a small number of public benchmarks compounds this through benchmark saturation, where methods are implicitly tuned to familiar test distributions. Patient-level data splits, explicit pretraining--test de-duplication, and rotating or held-out evaluation sets are needed for trustworthy comparison.

\textbf{Fairness, calibration, and uncertainty.} Downstream evaluation is still dominated by average accuracy or Dice, which can hide unequal performance across demographic or acquisition subgroups and says nothing about whether predicted confidences are trustworthy. For clinical deployment, \gls{ssl}-pretrained models should additionally be assessed for subgroup generalization, calibration, and reliable uncertainty estimates, since a well-aligned representation is of limited value if it fails silently on under-represented populations.

\textbf{Data governance and continual adaptation.} The scale of unlabeled data that makes \gls{ssl} attractive also raises privacy, licensing, and governance constraints, since large multi-institution corpora often cannot be centralized or redistributed. Privacy-preserving and federated pretraining offer a route to multi-site scale without direct data sharing, but require careful attention to consent and licensing provenance. Once deployed, models must also be updated as scanners, protocols, and populations drift, so continual adaptation without catastrophic forgetting or silent degradation is an open requirement.

\section{Future research directions}
\label{sec:future-research-directions}

We group promising directions into five themes, ordered from near-term priorities to more speculative opportunities.

\textbf{Pathology-aware pretext tasks.} Most current pretext objectives are generic and do not explicitly target the subtle, sparse features that drive clinical diagnosis. Designing tasks that prioritize diagnostically meaningful patterns, together with augmentation and masking policies adapted to medical rather than natural image statistics, is the most direct route to reducing the negative transfer documented in Section~\ref{sec:task-alignment-analysis}.

\textbf{Resource-efficient \gls{ssl} for high-dimensional data.} Computational cost is a practical barrier for pretraining on volumetric and high-resolution medical data. Memory-efficient architectures, lightweight representation learning, optimized 3D training schemes, and sparsity- or saliency-aware methods that focus computation on informative anatomical regions would broaden access to \gls{ssl}.

\textbf{Multi-modal, cross-dimensional, and longitudinal \gls{ssl}.} Clinical information is distributed across modalities, imaging dimensions, and time. Beyond the cross-modal and 2D/3D approaches reviewed here, longitudinal pretext tasks that model how anatomy and disease evolve across follow-up scans are largely unexplored. \glspl{jepa} and world-model objectives, which predict future or unseen states in latent space rather than reconstructing pixels, are a natural next step: they could support disease-progression tracking, cardiac and ultrasound motion modeling, and anticipation of subsequent frames in surgical video. Realizing these directions will require longitudinal and 4D datasets and evaluation protocols that reward accurate temporal prediction.

\textbf{Robust and acquisition-invariant pretraining across institutions.} Because models pretrained on one source may degrade elsewhere, future work should target acquisition-invariant pretext tasks that emphasize anatomy over scanner- or protocol-specific artifacts, together with decentralized or federated pretraining frameworks that enable multi-institutional use without direct data sharing. Adaptive objective selection---meta-learning or dynamic objective weighting tuned to the target task---could further improve robustness across heterogeneous distributions.

\textbf{Standardized evaluation and clinical validation.} Because recent gains often coincide with more powerful backbones and fine-tuning strategies, the independent contribution of the \gls{ssl} objective is difficult to isolate. Standardized protocols with common backbones, baselines, and label-regime reporting are needed for fair comparison, and evaluation should extend beyond accuracy to robustness, out-of-distribution generalization, calibration, and interpretability, and ultimately to external, prospective, workflow-level validation.

\section{Conclusion}
\label{sec:conclusion}
This review analyzed \gls{ssl} in medical imaging through a single question: whether the learning signal a pretext objective imposes preserves the information its downstream clinical task requires. Reading $78$ studies through that lens yields one organizing conclusion: \gls{ssl} effectiveness is not a property of a method but of a three-way match among pretext objective, imaging modality, and downstream demand, and it is strongest where labels are scarcest.

The contribution of this framing is practical. It converts an otherwise flat catalog of methods into conditional guidance, which objective to reach for given a target, a modality, and a label budget and, equally, it predicts when \gls{ssl} will fail, since the same misalignment that explains negative transfer also explains why no single method generalizes across the field. Realizing this in practice now depends less on new pretext tasks than on the evaluation gaps this review surfaces: attributing gains to the objective rather than the backbone, testing across institutions and subgroups rather than single benchmarks, and reporting calibration and uncertainty alongside accuracy. Progress on these will determine whether task-aligned \gls{ssl} translates from research settings into dependable clinical practice.

\section*{Declaration on the Use of Generative AI}
During the preparation of this work, the authors used OpenAI's ChatGPT to improve the language, grammar, and readability of portions of the manuscript. After using these tools, the authors reviewed and edited the content as needed and take full responsibility for the content of the publication. These tools were not used for research ideation, methodological design, data analysis, or the interpretation of results, all of which are solely the work of the authors.

\bibliographystyle{unsrt}  
\bibliography{references}

@article{willemink2020preparing,
  title={Preparing medical imaging data for machine learning},
  author={Willemink, Martin J and Koszek, Wojciech A and Hardell, Cailin and Wu, Jie and Fleischmann, Dominik and Harvey, Hugh and Folio, Les R and Summers, Ronald M and Rubin, Daniel L and Lungren, Matthew P},
  journal={Radiology},
  volume={295},
  number={1},
  pages={4--15},
  year={2020},
  publisher={Radiological Society of North America}
}

@article{tajbakhsh2020embracing,
  title={Embracing imperfect datasets: A review of deep learning solutions for medical image segmentation},
  author={Tajbakhsh, Nima and Jeyaseelan, Laura and Li, Qian and Chiang, Jeffrey N and Wu, Zhihao and Ding, Xiaowei},
  journal={Medical image analysis},
  volume={63},
  pages={101693},
  year={2020},
  publisher={Elsevier}
}

@article{jing2020self,
  title={Self-supervised visual feature learning with deep neural networks: A survey},
  author={Jing, Longlong and Tian, Yingli},
  journal={IEEE transactions on pattern analysis and machine intelligence},
  volume={43},
  number={11},
  pages={4037--4058},
  year={2020},
  publisher={IEEE}
}

@article{zhou2021review,
  title={A review of deep learning in medical imaging: Imaging traits, technology trends, case studies with progress highlights, and future promises},
  author={Zhou, S Kevin and Greenspan, Hayit and Davatzikos, Christos and Duncan, James S and Van Ginneken, Bram and Madabhushi, Anant and Prince, Jerry L and Rueckert, Daniel and Summers, Ronald M},
  journal={Proceedings of the IEEE},
  volume={109},
  number={5},
  pages={820--838},
  year={2021},
  publisher={IEEE}
}

@inproceedings{he2022masked,
  title={Masked autoencoders are scalable vision learners},
  author={He, Kaiming and Chen, Xinlei and Xie, Saining and Li, Yanghao and Doll{\'a}r, Piotr and Girshick, Ross},
  booktitle={Proceedings of the IEEE/CVF conference on computer vision and pattern recognition},
  pages={16000--16009},
  year={2022}
}

@inproceedings{chen2020simple,
  title={A simple framework for contrastive learning of visual representations},
  author={Chen, Ting and Kornblith, Simon and Norouzi, Mohammad and Hinton, Geoffrey},
  booktitle={International conference on machine learning},
  pages={1597--1607},
  year={2020},
  organization={PmLR}
}

@article{gidaris2018unsupervised,
  title={Unsupervised representation learning by predicting image rotations},
  author={Gidaris, Spyros and Singh, Praveer and Komodakis, Nikos},
  journal={arXiv preprint arXiv:1803.07728},
  year={2018}
}

@inproceedings{noroozi2016unsupervised,
  title={Unsupervised learning of visual representations by solving jigsaw puzzles},
  author={Noroozi, Mehdi and Favaro, Paolo},
  booktitle={European conference on computer vision},
  pages={69--84},
  year={2016},
  organization={Springer}
}

@article{purushwalkam2020demystifying,
  title={Demystifying contrastive self-supervised learning: Invariances, augmentations and dataset biases},
  author={Purushwalkam, Senthil and Gupta, Abhinav},
  journal={Advances in Neural Information Processing Systems},
  volume={33},
  pages={3407--3418},
  year={2020}
}

@article{chartsias2017multimodal,
  title={Multimodal MR synthesis via modality-invariant latent representation},
  author={Chartsias, Agisilaos and Joyce, Thomas and Giuffrida, Mario Valerio and Tsaftaris, Sotirios A},
  journal={IEEE transactions on medical imaging},
  volume={37},
  number={3},
  pages={803--814},
  year={2017},
  publisher={IEEE}
}

@article{armanious2020medgan,
  title={MedGAN: Medical image translation using GANs},
  author={Armanious, Karim and Jiang, Chenming and Fischer, Marc and K{\"u}stner, Thomas and Hepp, Tobias and Nikolaou, Konstantin and Gatidis, Sergios and Yang, Bin},
  journal={Computerized medical imaging and graphics},
  volume={79},
  pages={101684},
  year={2020},
  publisher={Elsevier}
}

@article{shurrab2022self,
  title={Self-supervised learning methods and applications in medical imaging analysis: A survey},
  author={Shurrab, Saeed and Duwairi, Rehab},
  journal={PeerJ Computer Science},
  volume={8},
  pages={e1045},
  year={2022},
  publisher={PeerJ Inc.}
}

@inproceedings{vincent2008extracting,
  title={Extracting and composing robust features with denoising autoencoders},
  author={Vincent, Pascal and Larochelle, Hugo and Bengio, Yoshua and Manzagol, Pierre-Antoine},
  booktitle={Proceedings of the 25th international conference on Machine learning},
  pages={1096--1103},
  year={2008}
}

@article{taleb20203d,
  title={3d self-supervised methods for medical imaging},
  author={Taleb, Aiham and Loetzsch, Winfried and Danz, Noel and Severin, Julius and Gaertner, Thomas and Bergner, Benjamin and Lippert, Christoph},
  journal={Advances in neural information processing systems},
  volume={33},
  pages={18158--18172},
  year={2020}
}

@inproceedings{he2020momentum,
  title={Momentum contrast for unsupervised visual representation learning},
  author={He, Kaiming and Fan, Haoqi and Wu, Yuxin and Xie, Saining and Girshick, Ross},
  booktitle={Proceedings of the IEEE/CVF conference on computer vision and pattern recognition},
  pages={9729--9738},
  year={2020}
}

@inproceedings{wang2021dense,
  title={Dense contrastive learning for self-supervised visual pre-training},
  author={Wang, Xinlong and Zhang, Rufeng and Shen, Chunhua and Kong, Tao and Li, Lei},
  booktitle={Proceedings of the IEEE/CVF conference on computer vision and pattern recognition},
  pages={3024--3033},
  year={2021}
}

@article{zhang2025survey,
  title={A survey on self-supervised learning: Recent advances and open problems},
  author={Zhang, Jianping and Yang, Lei and Mohammadabadi, Seyed Mahmoud Sajjadi and Yan, Feng},
  journal={Neurocomputing},
  pages={131409},
  year={2025},
  publisher={Elsevier}
}

@article{giakoumoglou2024review,
  title={A Review on Discriminative Self-supervised Learning Methods in Computer Vision},
  author={Giakoumoglou, Nikolaos and Stathaki, Tania and Gkelias, Athanasios},
  journal={arXiv preprint arXiv:2405.04969},
  year={2024}
}

@article{zhu2020rubik,
  title={Rubik’s cube+: A self-supervised feature learning framework for 3d medical image analysis},
  author={Zhu, Jiuwen and Li, Yuexiang and Hu, Yifan and Ma, Kai and Zhou, S Kevin and Zheng, Yefeng},
  journal={Medical image analysis},
  volume={64},
  pages={101746},
  year={2020},
  publisher={Elsevier}
}

@inproceedings{azizi2021big,
  title={Big self-supervised models advance medical image classification},
  author={Azizi, Shekoofeh and Mustafa, Basil and Ryan, Fiona and Beaver, Zachary and Freyberg, Jan and Deaton, Jonathan and Loh, Aaron and Karthikesalingam, Alan and Kornblith, Simon and Chen, Ting and others},
  booktitle={Proceedings of the IEEE/CVF international conference on computer vision},
  pages={3478--3488},
  year={2021}
}

@article{ciga2022self,
  title={Self supervised contrastive learning for digital histopathology},
  author={Ciga, Ozan and Xu, Tony and Martel, Anne Louise},
  journal={Machine learning with applications},
  volume={7},
  pages={100198},
  year={2022},
  publisher={Elsevier}
}

@inproceedings{ali2023self,
  title={Self-supervised learning for accurate liver view classification in ultrasound images with minimal labeled data},
  author={Ali, Abder-Rahman and Samir, Anthony E and Guo, Peng},
  booktitle={proceedings of the IEEE/CVF Conference on Computer Vision and Pattern Recognition},
  pages={3087--3093},
  year={2023}
}

@article{sriram2021covid,
  title={Covid-19 prognosis via self-supervised representation learning and multi-image prediction},
  author={Sriram, Anuroop and Muckley, Matthew and Sinha, Koustuv and Shamout, Farah and Pineau, Joelle and Geras, Krzysztof J and Azour, Lea and Aphinyanaphongs, Yindalon and Yakubova, Nafissa and Moore, William},
  journal={arXiv preprint arXiv:2101.04909},
  year={2021}
}

@article{chen2021momentum,
  title={Momentum contrastive learning for few-shot COVID-19 diagnosis from chest CT images},
  author={Chen, Xiaocong and Yao, Lina and Zhou, Tao and Dong, Jinming and Zhang, Yu},
  journal={Pattern recognition},
  volume={113},
  pages={107826},
  year={2021},
  publisher={Elsevier}
}

@article{cho2023chess,
  title={Chess: Chest x-ray pre-trained model via self-supervised contrastive learning},
  author={Cho, Kyungjin and Kim, Ki Duk and Nam, Yujin and Jeong, Jiheon and Kim, Jeeyoung and Choi, Changyong and Lee, Soyoung and Lee, Jun Soo and Woo, Seoyeon and Hong, Gil-Sun and others},
  journal={Journal of Digital Imaging},
  volume={36},
  number={3},
  pages={902--910},
  year={2023},
  publisher={Springer}
}

@article{nguyen2023self,
  title={Self-FI: self-supervised learning for disease diagnosis in fundus images},
  author={Nguyen, Toan Duc and Le, Duc-Tai and Bum, Junghyun and Kim, Seongho and Song, Su Jeong and Choo, Hyunseung},
  journal={Bioengineering},
  volume={10},
  number={9},
  pages={1089},
  year={2023},
  publisher={MDPI}
}

@article{li2020self,
  title={Self-supervised feature learning via exploiting multi-modal data for retinal disease diagnosis},
  author={Li, Xiaomeng and Jia, Mengyu and Islam, Md Tauhidul and Yu, Lequan and Xing, Lei},
  journal={IEEE Transactions on Medical Imaging},
  volume={39},
  number={12},
  pages={4023--4033},
  year={2020},
  publisher={IEEE}
}

@inproceedings{goncharov2023vox2vec,
  title={vox2vec: A framework for self-supervised contrastive learning of voxel-level representations in medical images},
  author={Goncharov, Mikhail and Soboleva, Vera and Kurmukov, Anvar and Pisov, Maxim and Belyaev, Mikhail},
  booktitle={International Conference on Medical Image Computing and Computer-Assisted Intervention},
  pages={605--614},
  year={2023},
  organization={Springer}
}

@article{ma2025multi,
  title={Multi-Encoding Contrastive Learning for Dual-Stream Self-Supervised 3D Dental Segmentation Network},
  author={Ma, Tian and Wei, Xiaoyuan and Zhai, Jiechen and Zhang, Ziang and Li, Yawen and Li, Yuancheng},
  journal={Technologies},
  volume={13},
  number={9},
  pages={419},
  year={2025},
  publisher={MDPI}
}

@article{yan2022deep,
  title={Deep contrastive learning based tissue clustering for annotation-free histopathology image analysis},
  author={Yan, Jiangpeng and Chen, Hanbo and Li, Xiu and Yao, Jianhua},
  journal={Computerized Medical Imaging and Graphics},
  volume={97},
  pages={102053},
  year={2022},
  publisher={Elsevier}
}

@inproceedings{vanberlo2023exploring,
  title={Exploring the utility of self-supervised pretraining strategies for the detection of absent lung sliding in m-mode lung ultrasound},
  author={VanBerlo, Blake and Li, Brian and Wong, Alexander and Hoey, Jesse and Arntfield, Robert},
  booktitle={Proceedings of the IEEE/CVF Conference on Computer Vision and Pattern Recognition},
  pages={3077--3086},
  year={2023}
}

@article{mh2023lvm,
  title={Lvm-med: Learning large-scale self-supervised vision models for medical imaging via second-order graph matching},
  author={MH Nguyen, Duy and Nguyen, Hoang and Diep, Nghiem and Pham, Tan Ngoc and Cao, Tri and Nguyen, Binh and Swoboda, Paul and Ho, Nhat and Albarqouni, Shadi and Xie, Pengtao and others},
  journal={Advances in Neural Information Processing Systems},
  volume={36},
  pages={27922--27950},
  year={2023}
}

@inproceedings{jamaludin2017self,
  title={Self-supervised learning for spinal MRIs},
  author={Jamaludin, Amir and Kadir, Timor and Zisserman, Andrew},
  booktitle={International Workshop on Deep Learning in Medical Image Analysis},
  pages={294--302},
  year={2017},
  organization={Springer}
}

@article{mohammad2024artificial,
  title={Artificial intelligence for detection of external cervical resorption using label-efficient self-supervised learning method},
  author={Mohammad-Rahimi, Hossein and Dianat, Omid and Abbasi, Reza and Zahedrozegar, Samira and Ashkan, Ali and Motamedian, Saeed Reza and Rohban, Mohammad Hossein and Nosrat, Ali},
  journal={Journal of Endodontics},
  volume={50},
  number={2},
  pages={144--153},
  year={2024},
  publisher={Elsevier}
}

@article{hu2025self,
  title={Self-supervised learning enhances periapical films segmentation with limited labeled data},
  author={Hu, Meiyu and Zhang, Qianli and Wei, Zhenyang and Jia, Pingyi and Yuan, Mu and Yu, Huajie and Yin, Xu-Cheng and Peng, Junran},
  journal={Journal of Dentistry},
  pages={106150},
  year={2025},
  publisher={Elsevier}
}

@article{xie2020pgl,
  title={PGL: Prior-guided local self-supervised learning for 3D medical image segmentation},
  author={Xie, Yutong and Zhang, Jianpeng and Liao, Zehui and Xia, Yong and Shen, Chunhua},
  journal={arXiv preprint arXiv:2011.12640},
  year={2020}
}

@inproceedings{hirsch2023self,
  title={Self-supervised learning for endoscopic video analysis},
  author={Hirsch, Roy and Caron, Mathilde and Cohen, Regev and Livne, Amir and Shapiro, Ron and Golany, Tomer and Goldenberg, Roman and Freedman, Daniel and Rivlin, Ehud},
  booktitle={International conference on medical image Computing and computer-assisted intervention},
  pages={569--578},
  year={2023},
  organization={Springer}
}

@article{taleb2022self,
  title={Self-supervised learning methods for label-efficient dental caries classification},
  author={Taleb, Aiham and Rohrer, Csaba and Bergner, Benjamin and De Leon, Guilherme and Rodrigues, Jonas Almeida and Schwendicke, Falk and Lippert, Christoph and Krois, Joachim},
  journal={Diagnostics},
  volume={12},
  number={5},
  pages={1237},
  year={2022},
  publisher={Multidisciplinary Digital Publishing Institute}
}

@inproceedings{zhuang2019self,
  title={Self-supervised feature learning for 3d medical images by playing a rubik’s cube},
  author={Zhuang, Xinrui and Li, Yuexiang and Hu, Yifan and Ma, Kai and Yang, Yujiu and Zheng, Yefeng},
  booktitle={International conference on medical image computing and computer-assisted intervention},
  pages={420--428},
  year={2019},
  organization={Springer}
}

@inproceedings{tao2020revisiting,
  title={Revisiting Rubik’s cube: Self-supervised learning with volume-wise transformation for 3D medical image segmentation},
  author={Tao, Xing and Li, Yuexiang and Zhou, Wenhui and Ma, Kai and Zheng, Yefeng},
  booktitle={International conference on medical image computing and computer-assisted intervention},
  pages={238--248},
  year={2020},
  organization={Springer}
}

@inproceedings{zhang2017self,
  title={Self supervised deep representation learning for fine-grained body part recognition},
  author={Zhang, Pengyue and Wang, Fusheng and Zheng, Yefeng},
  booktitle={2017 IEEE 14th international symposium on biomedical imaging (ISBI 2017)},
  pages={578--582},
  year={2017},
  organization={IEEE}
}

@article{nguyen2020self,
  title={Self-supervised learning based on spatial awareness for medical image analysis},
  author={Nguyen, Xuan-Bac and Lee, Guee Sang and Kim, Soo Hyung and Yang, Hyung Jeong},
  journal={IEEE access},
  volume={8},
  pages={162973--162981},
  year={2020},
  publisher={IEEE}
}

@inproceedings{bai2019self,
  title={Self-supervised learning for cardiac mr image segmentation by anatomical position prediction},
  author={Bai, Wenjia and Chen, Chen and Tarroni, Giacomo and Duan, Jinming and Guitton, Florian and Petersen, Steffen E and Guo, Yike and Matthews, Paul M and Rueckert, Daniel},
  booktitle={International conference on medical image computing and computer-assisted intervention},
  pages={541--549},
  year={2019},
  organization={Springer}
}

@inproceedings{spitzer2018improving,
  title={Improving cytoarchitectonic segmentation of human brain areas with self-supervised siamese networks},
  author={Spitzer, Hannah and Kiwitz, Kai and Amunts, Katrin and Harmeling, Stefan and Dickscheid, Timo},
  booktitle={International conference on medical image computing and computer-assisted intervention},
  pages={663--671},
  year={2018},
  organization={Springer}
}

@inproceedings{jiao2020self,
  title={Self-supervised representation learning for ultrasound video},
  author={Jiao, Jianbo and Droste, Richard and Drukker, Lior and Papageorghiou, Aris T and Noble, J Alison},
  booktitle={2020 IEEE 17th international symposium on biomedical imaging (ISBI)},
  pages={1847--1850},
  year={2020},
  organization={IEEE}
}

@article{manna2023self,
  title={Self-supervised representation learning for knee injury diagnosis from magnetic resonance data},
  author={Manna, Siladittya and Bhattacharya, Saumik and Pal, Umapada},
  journal={IEEE Transactions on Artificial Intelligence},
  volume={5},
  number={4},
  pages={1613--1623},
  year={2023},
  publisher={IEEE}
}

@inproceedings{taleb2021multimodal,
  title={Multimodal self-supervised learning for medical image analysis},
  author={Taleb, Aiham and Lippert, Christoph and Klein, Tassilo and Nabi, Moin},
  booktitle={International conference on information processing in medical imaging},
  pages={661--673},
  year={2021},
  organization={Springer}
}

@article{dong2023swssl,
  title={SWSSL: Sliding window-based self-supervised learning for anomaly detection in high-resolution images},
  author={Dong, Haoyu and Zhang, Yifan and Gu, Hanxue and Konz, Nicholas and Zhang, Yixin and Mazurowski, Maciej A},
  journal={IEEE Transactions on Medical Imaging},
  volume={42},
  number={12},
  pages={3860--3870},
  year={2023},
  publisher={IEEE}
}

@article{xing2023self,
  title={Self-supervised learning application on covid-19 chest x-ray image classification using masked autoencoder},
  author={Xing, Xin and Liang, Gongbo and Wang, Chris and Jacobs, Nathan and Lin, Ai-Ling},
  journal={Bioengineering},
  volume={10},
  number={8},
  pages={901},
  year={2023},
  publisher={MDPI}
}

@inproceedings{almalki2023self,
  title={Self-supervised learning with masked image modeling for teeth numbering, detection of dental restorations, and instance segmentation in dental panoramic radiographs},
  author={Almalki, Amani and Latecki, Longin Jan},
  booktitle={Proceedings of the IEEE/CVF winter conference on applications of computer vision},
  pages={5594--5603},
  year={2023}
}

@article{wang2023pyramid,
  title={Pyramid-based self-supervised learning for histopathological image classification},
  author={Wang, Junjie and Quan, Hao and Wang, Chengguang and Yang, Genke},
  journal={Computers in Biology and Medicine},
  volume={165},
  pages={107336},
  year={2023},
  publisher={Elsevier}
}

@article{wolf2023self,
  title={Self-supervised pre-training with contrastive and masked autoencoder methods for dealing with small datasets in deep learning for medical imaging},
  author={Wolf, Daniel and Payer, Tristan and Lisson, Catharina Silvia and Lisson, Christoph Gerhard and Beer, Meinrad and G{\"o}tz, Michael and Ropinski, Timo},
  journal={Scientific Reports},
  volume={13},
  number={1},
  pages={20260},
  year={2023},
  publisher={Nature Publishing Group UK London}
}

@inproceedings{almalki2024self,
  title={Self-supervised learning with masked autoencoders for teeth segmentation from intra-oral 3d scans},
  author={Almalki, Amani and Latecki, Longin Jan},
  booktitle={Proceedings of the IEEE/CVF Winter Conference on Applications of Computer Vision},
  pages={7820--7830},
  year={2024}
}

@inproceedings{zhou2019models,
  title={Models genesis: Generic autodidactic models for 3d medical image analysis},
  author={Zhou, Zongwei and Sodha, Vatsal and Rahman Siddiquee, Md Mahfuzur and Feng, Ruibin and Tajbakhsh, Nima and Gotway, Michael B and Liang, Jianming},
  booktitle={International conference on medical image computing and computer-assisted intervention},
  pages={384--393},
  year={2019},
  organization={Springer}
}

@article{chen2019self,
  title={Self-supervised learning for medical image analysis using image context restoration},
  author={Chen, Liang and Bentley, Paul and Mori, Kensaku and Misawa, Kazunari and Fujiwara, Michitaka and Rueckert, Daniel},
  journal={Medical image analysis},
  volume={58},
  pages={101539},
  year={2019},
  publisher={Elsevier}
}

@article{dominic2023improving,
  title={Improving data-efficiency and robustness of medical imaging segmentation using inpainting-based self-supervised learning},
  author={Dominic, Jeffrey and Bhaskhar, Nandita and Desai, Arjun D and Schmidt, Andrew and Rubin, Elka and Gunel, Beliz and Gold, Garry E and Hargreaves, Brian A and Lenchik, Leon and Boutin, Robert and others},
  journal={Bioengineering},
  volume={10},
  number={2},
  pages={207},
  year={2023},
  publisher={MDPI}
}

@inproceedings{patel2023self,
  title={Self-supervised anomaly detection from anomalous training data via iterative latent token masking},
  author={Patel, Ashay and Tudosiu, Petru-Daniel and Pinaya, Walter HL and Graham, Mark S and Adeleke, Olusola and Cook, Gary and Goh, Vicky and Ourselin, Sebastien and Cardoso, M Jorge},
  booktitle={Proceedings of the IEEE/CVF International Conference on Computer Vision},
  pages={2402--2410},
  year={2023}
}

@article{hervella2020learning,
  title={Learning the retinal anatomy from scarce annotated data using self-supervised multimodal reconstruction},
  author={Hervella, {\'A}lvaro S and Rouco, Jos{\'e} and Novo, Jorge and Ortega, Marcos},
  journal={Applied Soft Computing},
  volume={91},
  pages={106210},
  year={2020},
  publisher={Elsevier}
}

@article{morano2020multimodal,
  title={Multimodal transfer learning-based approaches for retinal vascular segmentation},
  author={Morano, Jos{\'e} and Hervella, {\'A}lvaro S and Barreira, Noelia and Novo, Jorge and Rouco, Jos{\'e}},
  journal={arXiv preprint arXiv:2012.10160},
  year={2020}
}

@article{hervella2021self,
  title={Self-supervised multimodal reconstruction pre-training for retinal computer-aided diagnosis},
  author={Hervella, {\'A}lvaro S and Rouco, Jose and Novo, Jorge and Ortega, Marcos},
  journal={Expert Systems with Applications},
  volume={185},
  pages={115598},
  year={2021},
  publisher={Elsevier}
}

@article{hervella2022multimodal,
  title={Multimodal image encoding pre-training for diabetic retinopathy grading},
  author={Hervella, Alvaro S and Rouco, Jos{\'e} and Novo, Jorge and Ortega, Marcos},
  journal={Computers in Biology and Medicine},
  volume={143},
  pages={105302},
  year={2022},
  publisher={Elsevier}
}

@inproceedings{cao2020auto,
  title={Auto-GAN: self-supervised collaborative learning for medical image synthesis},
  author={Cao, Bing and Zhang, Han and Wang, Nannan and Gao, Xinbo and Shen, Dinggang},
  booktitle={Proceedings of the AAAI conference on artificial intelligence},
  volume={34},
  number={07},
  pages={10486--10493},
  year={2020}
}

@article{ross2018exploiting,
  title={Exploiting the potential of unlabeled endoscopic video data with self-supervised learning},
  author={Ross, Tobias and Zimmerer, David and Vemuri, Anant and Isensee, Fabian and Wiesenfarth, Manuel and Bodenstedt, Sebastian and Both, Fabian and Kessler, Philip and Wagner, Martin and M{\"u}ller, Beat and others},
  journal={International journal of computer assisted radiology and surgery},
  volume={13},
  number={6},
  pages={925--933},
  year={2018},
  publisher={Springer}
}

@article{haghighi2024self,
  title={Self-supervised learning for medical image analysis: Discriminative, restorative, or adversarial?},
  author={Haghighi, Fatemeh and Taher, Mohammad Reza Hosseinzadeh and Gotway, Michael B and Liang, Jianming},
  journal={Medical Image Analysis},
  volume={94},
  pages={103086},
  year={2024},
  publisher={Elsevier}
}

@article{qayyum2023two,
  title={Two-stage self-supervised contrastive learning aided transformer for real-time medical image segmentation},
  author={Qayyum, Abdul and Razzak, Imran and Mazher, Moona and Khan, Tariq and Ding, Weiping and Niederer, Steven},
  journal={IEEE Journal of Biomedical and Health Informatics},
  year={2023},
  publisher={IEEE}
}

@inproceedings{dong2021self,
  title={Self-supervised multi-task representation learning for sequential medical images},
  author={Dong, Nanqing and Kampffmeyer, Michael and Voiculescu, Irina},
  booktitle={Joint European Conference on Machine Learning and Knowledge Discovery in Databases},
  pages={779--794},
  year={2021},
  organization={Springer}
}

@article{li2021rotation,
  title={Rotation-oriented collaborative self-supervised learning for retinal disease diagnosis},
  author={Li, Xiaomeng and Hu, Xiaowei and Qi, Xiaojuan and Yu, Lequan and Zhao, Wei and Heng, Pheng-Ann and Xing, Lei},
  journal={IEEE Transactions on Medical Imaging},
  volume={40},
  number={9},
  pages={2284--2294},
  year={2021},
  publisher={IEEE}
}

@article{song2022covid,
  title={COVID-19 infection segmentation and severity assessment using a self-supervised learning approach},
  author={Song, Yao and Liu, Jun and Liu, Xinghua and Tang, Jinshan},
  journal={Diagnostics},
  volume={12},
  number={8},
  pages={1805},
  year={2022},
  publisher={MDPI}
}

@article{lu2021volumetric,
  title={Volumetric white matter tract segmentation with nested self-supervised learning using sequential pretext tasks},
  author={Lu, Qi and Li, Yuxing and Ye, Chuyang},
  journal={Medical Image Analysis},
  volume={72},
  pages={102094},
  year={2021},
  publisher={Elsevier}
}

@inproceedings{tajbakhsh2019surrogate,
  title={Surrogate supervision for medical image analysis: Effective deep learning from limited quantities of labeled data},
  author={Tajbakhsh, Nima and Hu, Yufei and Cao, Junli and Yan, Xingjian and Xiao, Yi and Lu, Yong and Liang, Jianming and Terzopoulos, Demetri and Ding, Xiaowei},
  booktitle={2019 IEEE 16th international symposium on biomedical imaging (ISBI 2019)},
  pages={1251--1255},
  year={2019},
  organization={IEEE}
}

@article{koohbanani2021self,
  title={Self-path: Self-supervision for classification of pathology images with limited annotations},
  author={Koohbanani, Navid Alemi and Unnikrishnan, Balagopal and Khurram, Syed Ali and Krishnaswamy, Pavitra and Rajpoot, Nasir},
  journal={IEEE Transactions on Medical Imaging},
  volume={40},
  number={10},
  pages={2845--2856},
  year={2021},
  publisher={IEEE}
}

@article{luo2020retinal,
  title={Retinal image classification by self-supervised fuzzy clustering network},
  author={Luo, Yueguo and Pan, Jing and Fan, Shaoshuai and Du, Zeyu and Zhang, Guanghua},
  journal={IEEE Access},
  volume={8},
  pages={92352--92362},
  year={2020},
  publisher={IEEE}
}

@inproceedings{li2020multi,
  title={A multi-task self-supervised learning framework for scopy images},
  author={Li, Yuexiang and Chen, Jiawei and Zheng, Yefeng},
  booktitle={2020 IEEE 17th international symposium on biomedical imaging (ISBI)},
  pages={2005--2009},
  year={2020},
  organization={IEEE}
}

@article{xie2024unimiss+,
  title={UniMiSS+: Universal medical self-supervised learning from cross-dimensional unpaired data},
  author={Xie, Yutong and Zhang, Jianpeng and Xia, Yong and Wu, Qi},
  journal={IEEE Transactions on Pattern Analysis and Machine Intelligence},
  volume={46},
  number={12},
  pages={10021--10035},
  year={2024},
  publisher={IEEE}
}

@article{wang2024multi,
  title={A multi-resolution self-supervised learning framework for semantic segmentation in histopathology},
  author={Wang, Hao and Ahn, Euijoon and Kim, Jinman},
  journal={Pattern Recognition},
  volume={155},
  pages={110621},
  year={2024},
  publisher={Elsevier}
}

@inproceedings{yi2024parcl,
  title={PaRCL: Pathology-aware Representation Contrastive Learning for Glaucoma Classification on Fundus Images},
  author={Yi, Junyan and Zheng, Ying and Ding, Dayong and Zhao, Jianchun and Yang, Gang},
  booktitle={2024 IEEE International Conference on Bioinformatics and Biomedicine (BIBM)},
  pages={3912--3917},
  year={2024},
  organization={IEEE}
}

@inproceedings{daydar2024med,
  title={Med-SeAM: medical context aware self-supervised learning framework for anomaly classification in knee MRI},
  author={Daydar, Akshay and Reddy, Ajay Kumar and Kumar, Sonal and Sur, Arijit and Laskar, Hanif},
  booktitle={Proceedings of the Fifteenth Indian Conference on Computer Vision Graphics and Image Processing},
  pages={1--8},
  year={2024}
}

@inproceedings{nasiri2024vim4path,
  title={Vim4path: Self-supervised vision mamba for histopathology images},
  author={Nasiri-Sarvi, Ali and Trinh, Vincent Quoc-Huy and Rivaz, Hassan and Hosseini, Mahdi S},
  booktitle={Proceedings of the IEEE/CVF conference on computer vision and pattern recognition},
  pages={6894--6903},
  year={2024}
}

@article{ye2024cads,
  title={CADS: A self-supervised learner via cross-modal alignment and deep self-distillation for CT volume segmentation},
  author={Ye, Yiwen and Zhang, Jianpeng and Chen, Ziyang and Xia, Yong},
  journal={IEEE Transactions on Medical Imaging},
  volume={44},
  number={1},
  pages={118--129},
  year={2024},
  publisher={IEEE}
}

@inproceedings{tahghighi2024enhancing,
  title={Enhancing new multiple sclerosis lesion segmentation via self-supervised pre-training and synthetic lesion integration},
  author={Tahghighi, Peyman and Zhang, Yunyan and Souza, Roberto and Komeili, Amin},
  booktitle={International Conference on Medical Image Computing and Computer-Assisted Intervention},
  pages={263--272},
  year={2024},
  organization={Springer}
}

@article{ozbay2024kidney,
  title={Kidney tumor classification on CT images using self-supervised learning},
  author={{\"O}zbay, Erdal and {\"O}zbay, Feyza Altunbey and Gharehchopogh, Farhad Soleimanian},
  journal={Computers in Biology and Medicine},
  volume={176},
  pages={108554},
  year={2024},
  publisher={Elsevier}
}

@article{zhao2025maemc,
  title={MAEMC-NET: a hybrid self-supervised learning method for predicting the malignancy of solitary pulmonary nodules from CT images},
  author={Zhao, Tianhu and Yue, Yong and Sun, Hang and Li, Jingxu and Wen, Yanhua and Yao, Yudong and Qian, Wei and Guan, Yubao and Qi, Shouliang},
  journal={Frontiers in medicine},
  volume={12},
  pages={1507258},
  year={2025},
  publisher={Frontiers Media SA}
}

@article{de2025self,
  title={Self-supervised learning leads to improved performance in biparametric prostate MRI classification},
  author={de Almeida, Jos{\'e} Guilherme and Verde, Ana Sofia Castro and Gaiv{\~a}o, Ana Mascarenhas and Bilreiro, Carlos and Santiago, In{\^e}s and Ip, Joana and Beli{\~a}o, Sara and Matos, Celso and Tsiknakis, Manolis and Marias, Kostas and others},
  journal={Computers in Biology and Medicine},
  volume={198},
  pages={111262},
  year={2025},
  publisher={Elsevier}
}

@article{meng2024self,
  title={Self-supervised contrastive learning for automated segmentation of brain tumor MRI images in schizophrenia},
  author={Meng, Lingmiao and Zhao, Liwei and Yi, Xin and Yu, Qingming},
  journal={International Journal of Computational Intelligence Systems},
  volume={17},
  number={1},
  pages={196},
  year={2024},
  publisher={Springer}
}

@article{gryshchuk2025contrastive,
  title={Contrastive self-supervised learning for neurodegenerative disorder classification},
  author = {Gryshchuk, Vadym and Singh, Devesh and Teipel, Stefan and
          Dyrba, Martin and {{ADNI, AIBL, and FTLDNI Study Groups}}},
  journal={Frontiers in Neuroinformatics},
  volume={19},
  pages={1527582},
  year={2025},
  publisher={Frontiers Media SA}
}

@article{suekei2024multi,
  title={Multi-modal representation learning in retinal imaging using self-supervised learning for enhanced clinical predictions},
  author={Suekei, Emese and Rumetshofer, Elisabeth and Schmidinger, Niklas and Mayr, Andreas and Schmidt-Erfurth, Ursula and Klambauer, Guenter and Bogunovi{\'c}, Hrvoje},
  journal={Scientific Reports},
  volume={14},
  number={1},
  pages={26802},
  year={2024},
  publisher={Nature Publishing Group UK London}
}

@article{ekanayake2025cl,
  title={CL-MRI: Self-Supervised contrastive learning to improve the accuracy of undersampled MRI reconstruction},
  author={Ekanayake, Mevan and Chen, Zhifeng and Harandi, Mehrtash and Egan, Gary and Chen, Zhaolin},
  journal={Biomedical Signal Processing and Control},
  volume={100},
  pages={107185},
  year={2025},
  publisher={Elsevier}
}

@inproceedings{moreau2025self,
  title={Self-supervised learning for stroke lesion segmentation on CT: a new pretext task for neuroimaging},
  author={Moreau, Juliette and Mechtouff, Laura and Rousseau, David and Cho, Tae-Hee and Eker, Omer and Berthez{\'e}ne, Yves and Frindel, Carole},
  booktitle={2025 47th Annual International Conference of the IEEE Engineering in Medicine and Biology Society (EMBC)},
  pages={1--6},
  year={2025},
  organization={IEEE}
}

@article{liu2025perfusion,
  title={Perfusion estimation from dynamic non-contrast computed tomography using self-supervised learning and a physics-inspired U-net transformer architecture},
  author={Liu, Yi-Kuan and Cisneros, Jorge and Nair, Girish and Stevens, Craig and Castillo, Richard and Vinogradskiy, Yevgeniy and Castillo, Edward},
  journal={International Journal of Computer Assisted Radiology and Surgery},
  volume={20},
  number={5},
  pages={959--970},
  year={2025},
  publisher={Springer}
}

@article{chyrmang2025self,
  title={Self-HER2Net: A generative self-supervised framework for HER2 classification in IHC histopathology of breast cancer},
  author={Chyrmang, Genevieve and Barua, Barun and Bora, Kangkana and Ahmed, Gazi N and Das, Anup Kr and Kakoti, Lopamudra and Lemos, Bernardo and Mallik, Saurav},
  journal={Pathology-Research and Practice},
  volume={270},
  pages={155961},
  year={2025},
  publisher={Elsevier}
}

@article{khalid2025improving,
  title={Improving Coronary Artery Disease Diagnosis in Cardiac MRI with Self-Supervised Learning},
  author={Khalid, Usman and Kaya, Mehmet and Alhajj, Reda},
  journal={Diagnostics},
  volume={15},
  number={20},
  pages={2618},
  year={2025},
  publisher={MDPI}
}

@article{wang2025self,
  title={Self-supervised graph contrastive learning with diffusion augmentation for functional MRI analysis and brain disorder detection},
  author={Wang, Xiaochuan and Fang, Yuqi and Wang, Qianqian and Yap, Pew-Thian and Zhu, Hongtu and Liu, Mingxia},
  journal={Medical image analysis},
  volume={101},
  pages={103403},
  year={2025},
  publisher={Elsevier}
}

@article{liu20243d,
  title={3D segmentation of necrotic lung lesions in CT images using self-supervised contrastive learning},
  author={Liu, Yiqiao and Halek, Sarah and Crawford, Randolph and Persson, Keith and Tomaszewski, Michal and Wang, Shubing and Baumgartner, Richard and Yuan, Jianda and Goldmacher, Gregory and Chen, Antong},
  journal={IEEE Access},
  volume={12},
  pages={32859--32869},
  year={2024},
  publisher={IEEE}
}

@article{huang2023ssl,
  title={Self-supervised learning for medical image classification: a systematic review and implementation guidelines},
  author={Huang, Shih-Cheng and Pareek, Anuj and Jensen, Malte and Lungren, Matthew P and Yeung, Serena and Chaudhari, Akshay S},
  journal={NPJ Digital Medicine},
  volume={6},
  number={1},
  pages={74},
  year={2023},
  publisher={Nature Publishing Group UK London}
}

@article{vanberlo2024survey,
  title={A survey of the impact of self-supervised pretraining for diagnostic tasks in medical X-ray, CT, MRI, and ultrasound},
  author={VanBerlo, Blake and Hoey, Jesse and Wong, Alexander},
  journal={BMC medical imaging},
  volume={24},
  number={1},
  pages={79},
  year={2024},
  publisher={Springer}
}

@article{zhang2023dive,
  title={Dive into the details of self-supervised learning for medical image analysis},
  author={Zhang, Chuyan and Zheng, Hao and Gu, Yun},
  journal={Medical Image Analysis},
  volume={89},
  pages={102879},
  year={2023},
  publisher={Elsevier}
}

@article{zeng2024self,
  title={Self-supervised learning framework application for medical image analysis: a review and summary},
  author={Zeng, Xiangrui and Abdullah, Nibras and Sumari, Putra},
  journal={BioMedical Engineering OnLine},
  volume={23},
  number={1},
  pages={107},
  year={2024},
  publisher={Springer}
}

@article{kumari2026learning,
  title={Learning with less: A survey of deep learning in medical imaging under varying supervision levels},
  author={Kumari, Suruchi and Singh, Pravendra},
  journal={Artificial Intelligence in Medicine},
  pages={103375},
  year={2026},
  publisher={Elsevier}
}

@inproceedings{wu2024voco,
  title={Voco: A simple-yet-effective volume contrastive learning framework for 3d medical image analysis},
  author={Wu, Linshan and Zhuang, Jiaxin and Chen, Hao},
  booktitle={Proceedings of the IEEE/CVF conference on computer vision and pattern recognition},
  pages={22873--22882},
  year={2024}
}

@article{zhou2023pcrlv2,
  title={A unified visual information preservation framework for self-supervised pre-training in medical image analysis},
  author={Zhou, Hong-Yu and Lu, Chixiang and Chen, Chaoqi and Yang, Sibei and Yu, Yizhou},
  journal={IEEE Transactions on Pattern Analysis and Machine Intelligence},
  volume={45},
  number={7},
  pages={8020--8035},
  year={2023},
  publisher={IEEE}
}

@inproceedings{tang2022self,
  title={Self-supervised pre-training of swin transformers for 3d medical image analysis},
  author={Tang, Yucheng and Yang, Dong and Li, Wenqi and Roth, Holger R and Landman, Bennett and Xu, Daguang and Nath, Vishwesh and Hatamizadeh, Ali},
  booktitle={Proceedings of the IEEE/CVF conference on computer vision and pattern recognition},
  pages={20730--20740},
  year={2022}
}

@inproceedings{wald2025revisiting,
  title={Revisiting MAE pre-training for 3D medical image segmentation},
  author={Wald, Tassilo and Ulrich, Constantin and Lukyanenko, Stanislav and Goncharov, Andrei and Paderno, Alberto and Miller, Maximilian and Maerkisch, Leander and Jaeger, Paul and Maier-Hein, Klaus},
  booktitle={Proceedings of the Computer Vision and Pattern Recognition Conference},
  pages={5186--5196},
  year={2025}
}

@article{wang2022ctranspath,
  title={Transformer-based unsupervised contrastive learning for histopathological image classification},
  author={Wang, Xiyue and Yang, Sen and Zhang, Jun and Wang, Minghui and Zhang, Jing and Yang, Wei and Huang, Junzhou and Han, Xiao},
  journal={Medical image analysis},
  volume={81},
  pages={102559},
  year={2022},
  publisher={Elsevier}
}

@inproceedings{chen2022hipt,
  title={Scaling vision transformers to gigapixel images via hierarchical self-supervised learning},
  author={Chen, Richard J and Chen, Chengkuan and Li, Yicong and Chen, Tiffany Y and Trister, Andrew D and Krishnan, Rahul G and Mahmood, Faisal},
  booktitle={Proceedings of the IEEE/CVF conference on computer vision and pattern recognition},
  pages={16144--16155},
  year={2022}
}

@article{vorontsov2023virchow,
  title={A foundation model for clinical-grade computational pathology and rare cancers detection},
  author={Vorontsov, Eugene and Bozkurt, Alican and Casson, Adam and Shaikovski, George and Zelechowski, Michal and Severson, Kristen and Zimmermann, Eric and Hall, James and Tenenholtz, Neil and Fusi, Nicolo and others},
  journal={Nature medicine},
  volume={30},
  number={10},
  pages={2924--2935},
  year={2024},
  publisher={Nature Publishing Group US New York}
}

@article{xu2024provgigapath,
  title={A whole-slide foundation model for digital pathology from real-world data},
  author={Xu, Hanwen and Usuyama, Naoto and Bagga, Jaspreet and Zhang, Sheng and Rao, Rajesh and Naumann, Tristan and Wong, Cliff and Gero, Zelalem and Gonz{\'a}lez, Javier and Gu, Yu and others},
  journal={Nature},
  volume={630},
  number={8015},
  pages={181--188},
  year={2024},
  publisher={Nature Publishing Group UK London}
}

@article{zhou2023retfound,
  title={A foundation model for generalizable disease detection from retinal images},
  author={Zhou, Yukun and Chia, Mark A and Wagner, Siegfried K and Ayhan, Murat S and Williamson, Dominic J and Struyven, Robbert R and Liu, Timing and Xu, Moucheng and Lozano, Mateo G and Woodward-Court, Peter and others},
  journal={Nature},
  volume={622},
  number={7981},
  pages={156--163},
  year={2023},
  publisher={Nature Publishing Group UK London}
}

@article{xu2025_3dino,
  title={A generalizable 3D framework and model for self-supervised learning in medical imaging},
  author={Xu, Tony and Hosseini, Sepehr and Anderson, Chris and Rinaldi, Anthony and Krishnan, Rahul G and Martel, Anne L and Goubran, Maged},
  journal={npj Digital Medicine},
  volume={8},
  number={1},
  pages={639},
  year={2025},
  publisher={Nature Publishing Group UK London}
}

@inproceedings{xie2022simmim,
  title={Simmim: A simple framework for masked image modeling},
  author={Xie, Zhenda and Zhang, Zheng and Cao, Yue and Lin, Yutong and Bao, Jianmin and Yao, Zhuliang and Dai, Qi and Hu, Han},
  booktitle={Proceedings of the IEEE/CVF conference on computer vision and pattern recognition},
  pages={9653--9663},
  year={2022}
}

@article{zhang2024anatomy,
  title={Self-supervised learning for medical image data with anatomy-oriented imaging planes},
  author={Zhang, Tianwei and Wei, Dong and Zhu, Mengmeng and Gu, Shi and Zheng, Yefeng},
  journal={Medical Image Analysis},
  volume={94},
  pages={103151},
  year={2024},
  publisher={Elsevier}
}

@article{grill2020byol,
  title={Bootstrap your own latent-a new approach to self-supervised learning},
  author={Grill, Jean-Bastien and Strub, Florian and Altch{\'e}, Florent and Tallec, Corentin and Richemond, Pierre and Buchatskaya, Elena and Doersch, Carl and Avila Pires, Bernardo and Guo, Zhaohan and Gheshlaghi Azar, Mohammad and others},
  journal={Advances in neural information processing systems},
  volume={33},
  pages={21271--21284},
  year={2020}
}

@inproceedings{chen2021simsiam,
  title={Exploring simple siamese representation learning},
  author={Chen, Xinlei and He, Kaiming},
  booktitle={Proceedings of the IEEE/CVF conference on computer vision and pattern recognition},
  pages={15750--15758},
  year={2021}
}

@inproceedings{caron2021dino,
  title={Emerging properties in self-supervised vision transformers},
  author={Caron, Mathilde and Touvron, Hugo and Misra, Ishan and J{\'e}gou, Herv{\'e} and Mairal, Julien and Bojanowski, Piotr and Joulin, Armand},
  booktitle={Proceedings of the IEEE/CVF international conference on computer vision},
  pages={9650--9660},
  year={2021}
}

@inproceedings{zbontar2021barlow,
  title={Barlow twins: Self-supervised learning via redundancy reduction},
  author={Zbontar, Jure and Jing, Li and Misra, Ishan and LeCun, Yann and Deny, St{\'e}phane},
  booktitle={International conference on machine learning},
  pages={12310--12320},
  year={2021},
  organization={PMLR}
}

@article{bardes2022vicreg,
  title={Vicreg: Variance-invariance-covariance regularization for self-supervised learning},
  author={Bardes, Adrien and Ponce, Jean and LeCun, Yann},
  journal={arXiv preprint arXiv:2105.04906},
  year={2021}
}

\end{document}